\documentclass[lettersize,journal]{IEEEtran}
\usepackage{amsmath,amsfonts}
\usepackage{algorithmic}
\usepackage{array}
\usepackage[caption=false,font=normalsize,labelfont=sf,textfont=sf]{subfig}
\usepackage[colorlinks=true, linkcolor=blue, citecolor=blue, urlcolor=blue, pdfborder={0 0 0}]{hyperref}

\usepackage{textcomp}
\usepackage{url}
\usepackage[caption=false,font=normalsize,labelfont=sf,textfont=sf]{subfig}
\usepackage{verbatim}
\usepackage{graphicx}
\usepackage{color}
\usepackage{xcolor}
\newtheorem{remark}{Remark}

\usepackage[linesnumbered, ruled,vlined]{algorithm2e}

\usepackage{booktabs}

\usepackage[switch]{lineno}

\usepackage[table]{xcolor}
\usepackage{multicol}
\usepackage{multirow}
\usepackage{pifont}
\usepackage{balance}
\usepackage{bbding}

\begin{document}

\title{AOI-Net: Structural Face AOI-Guided Eye-Gaze Track Representation Learning for Autism Spectrum Disorder Detection}

\author{
Zhanpei~Huang,
Binbin~Sun,
Jialiang~Chen,
Yiou~Wang,
Taochen~Chen,
Yuzhu~Ji,~\IEEEmembership{Senior Member,~IEEE},\\
Yiqun~Zhang,~\IEEEmembership{Senior Member,~IEEE},
and Yiu-Ming~Cheung,~\IEEEmembership{Fellow,~IEEE}
\thanks{Received 18 March 2026; revised 6 June 2026; revised 1 August 2026; accepted 24 August 2026. (Zhanpei Huang and Binbin Sun contributed equally to this work. Yiqun Zhang is the corresponding author.) }
\thanks{Zhanpei Huang, Jialiang Chen, Yiou Wang, Taochen Chen, Yuzhu Ji, and Yiqun Zhang are with the School of Computer Science and Technology, Guangdong University of Technology, Guangzhou, China (e-mail: \{2112405010, 3124004088, 3223004344\}@mail2.gdut.edu.cn, chentaochen1@mails.gdut.edu.cn, \{yuzhu.ji, yqzhang\}@gdut.edu.cn).}
\thanks{Binbin Sun is with the Shenzhen Maternity and Child Healthcare Hospital Affiliated to Southern Medical University, Shenzhen, China (e-mail: sunbin13530280620@126.com).}
\thanks{Yiu-Ming Cheung is with the Department of Computer Science, Hong Kong Baptist University, Hong Kong SAR, China (e-mail: ymc@comp.hkbu.edu.hk). }
}

\markboth{IEEE Computational Intelligence Magazine}{}

\maketitle

\begin{abstract}

Eye-movement tracking has emerged as a promising non-invasive approach to Autism Spectrum Disorder (ASD) screening, with systematic differences in attentional allocation and revisit behaviors observed during socially interactive tasks. Existing computational methods typically characterize eye-movements using discrete gaze trajectories and fixation events, yielding representations dominated by short-range temporal dynamics and limiting models that primarily emphasize long-range dependencies. Meanwhile, gaze behavior is naturally organized across semantically meaningful Areas of Interest (AOIs), whose attention allocation and transitions provide important structural cues, yet their relationships are rarely modeled explicitly. To address these limitations, we propose a structural face AOI-guided Eye-Gaze Track Network (AOI-Net) that jointly models short-term temporal dynamics and AOI-level structural organization. A network gating mechanism adaptively integrates the complementary temporal and structural representations according to their contributions to gaze-behavior characterization. To mitigate the pronounced class imbalance commonly encountered between individuals with ASD and Typically Developing (TD) participants in clinical datasets, class-distribution-aware learning is further employed to facilitate discriminative embedding learning under skewed class distributions. Experiments on a unique and large-scale clinical eye-tracking database comprising eight stimulus subsets and more than 1,300 participants show that AOI-Net consistently outperforms state-of-the-art methods. The proposed framework also enables interpretable gaze-behavior modeling and provides a practical basis for scalable AI-driven ASD screening in real-world healthcare. The code is available \href{https://github.com/Zhanpei-ai/CIM-AOI-Net/tree/main/Code}{here}.

\end{abstract}


\section{Introduction}
\IEEEPARstart{A}{utism} Spectrum Disorder (ASD) is a neurodevelopmental condition characterized by persistent difficulties in social interaction, communication, and attentional regulation~\cite{lord2018autism,lord2020autism,Ting2024Medicine}. Early identification of ASD is crucial for timely intervention, which has been shown to significantly improve long-term developmental outcomes~\cite{zwaigenbaum2015early,dawson2010randomized,eldevik2026clinically}. However, the existing clinical diagnostic procedures rely on expert observation and behavioral scales, which are time-consuming and difficult to access for large populations. 
That is, standardized diagnostic instruments such as the Autism Diagnostic Observation Schedule (ADOS)~\cite{gotham2007autism} and the Autism Diagnostic Interview-Revised (ADI-R)~\cite{rutter2003autism} rely on trained clinicians to conduct structured or semi-structured behavioral assessments and caregiver interviews under controlled settings to characterize social, communicative, and developmental patterns.
These limitations have attracted increasing attention, motivating the development of objective, data-driven methods for the automatic diagnosis of neurodevelopmental disorders~\cite{qureshi2020improving,ramirez2026artificial}, among which eye-movement–based approaches have shown particular promise for ASD screening~\cite{Zhang2025Uncertainty}.

\begin{figure}
    \centering
    \includegraphics[width=1\linewidth]{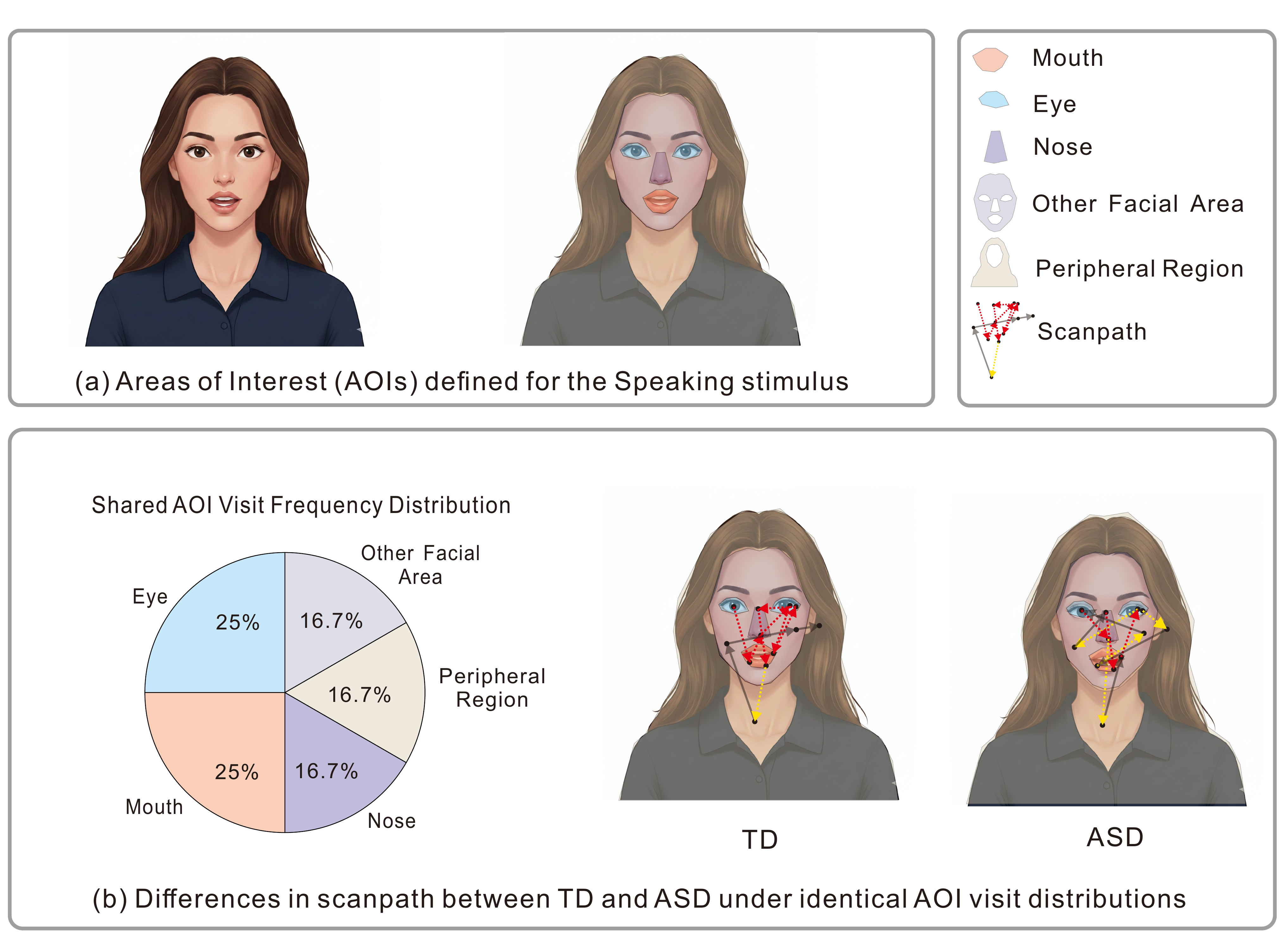}
    \caption{Schematic Illustration of the Eye-Tracking Stimulus Paradigm and AOI Statistics for ASD Screening. 
    Red dashed lines indicate transitions within social regions, and yellow dashed lines represent shifts from social to non-social regions. (Note: Due to copyright restrictions of the original visual stimuli, the woman's face shown in this figure is an AI-generated synthetic image used solely for illustrative purposes.)}
    \label{fig:aoi}

\end{figure}

Eye-movement behavior has been increasingly recognized in medical and cognitive research as an informative and non-invasive signal for ASD-related assessment~\cite{falck2013eye, billeci2016disentangling}. 
During interactive or socially rich video-based tasks, Typically Developing (TD) children tend to allocate visual attention in a coordinated manner toward socially relevant cues, whereas children with ASD often exhibit atypical attentional organization, characterized by reduced engagement with social regions and frequent shifts of visual focus~\cite{constantino2017infant,frazier2017meta}.
Eye-movement recordings collected during such tasks can therefore provide informative behavioral signals for ASD-related assessment. 
In dynamic visual scenes, eye-movements naturally form temporal sequences composed of discrete fixation events connected by rapid saccadic transitions, where each fixation typically falls on a semantically meaningful region of the scene, often defined as an Area of Interest (AOI), as illustrated in Fig.~\ref{fig:aoi}(a).

Early studies on ASD detection based on eye-movement data primarily focused on static feature-based representations. These approaches summarize eye-movement behavior using handcrafted statistical descriptors extracted from fixation and saccade events, such as fixation duration, fixation count, and velocity~\cite{xu2024portable, kong2022different}. By aggregating eye-movement recordings into fixed-length feature vectors, static feature-based methods enable the use of conventional classifiers for ASD identification~\cite{jones2013attention,constantino2017infant}. 
Although they capture global behavioral tendencies, they discard the temporal order of fixation events and overlook fixation-to-fixation transitions, limiting their ability to represent the sequential structure of gaze behavior.

To dynamically capture temporal dependencies in sequential data, general-purpose time-series approaches largely rely on global modeling techniques designed to model long-range temporal dependencies. For instance, recent time-series models, such as Informer~\cite{zhou2020informer}, TimesNet~\cite{timesnet}, MPTSNet~\cite{mu2025mptsnet}, and XPatch~\cite{xpatch}, focus on global temporal interactions over extended sequences. These methods are inherently developed for long-range sequences, implicitly assuming continuous temporal dynamics. In contrast, eye-movements elicited by dynamic social stimuli are event-driven in nature, consisting of discrete fixation events connected by rapid saccadic transitions.
Fixation-based eye-movement behavior is characterized by short-range temporal dependencies, as eye-movement control theories suggest that gaze behavior is primarily governed by local interactions between successive fixations~\cite{engbert2005swift}. Behaviorally meaningful gaze patterns primarily emerge from transitions between temporally adjacent fixations, rather than from associations between distant events. When applied to fixation-based eye-movement sequences, enforcing long-range temporal interactions may therefore weaken the modeling of behaviorally relevant patterns.
Associations between temporally distant fixations often lack direct attentional relevance, and explicitly modeling such relations can introduce unstable dependencies that are more susceptible to noise.
As a result, the model may allocate representational capacity to weak correlations, reducing its ability to capture informative local transitions that reflect meaningful gaze behavior.

Beyond temporal sequence modeling, another line of research incorporates semantic scene information through AOIs, which provide an interpretable representation of gaze behavior and are particularly valuable for trustworthy medical applications~\cite{Tjoa2021Survey}. 
Early AOI-based studies characterized gaze using statistical descriptors such as visitation frequency or fixation duration within predefined regions~\cite{wang2015atypical,frazier2017meta}. However, these global statistics ignore the sequential transitions between fixations and discard information about the ordering of attention. Even when two gaze recordings exhibit identical visitation frequencies across AOIs, the actual sequences of fixation transitions can differ substantially, as illustrated in Fig.~\ref{fig:aoi}(b).
More recent sequence modeling approaches, such as EmMixformer~\cite{qin2025emmixformer} and Detach\_ROCKET~\cite{uribarri2024detach} treat eye-movement data as unified temporal sequences and apply generic sequence-processing architectures. Although these methods capture temporal dynamics, AOI information is typically incorporated as categorical variables using one-hot encodings or learned embeddings. Such representations treat AOIs as independent labels and fail to capture the structural organization and relational dependencies among regions of interest.
Taken together, existing approaches either emphasize temporal sequence modeling while neglecting AOI structural information, or rely on AOI statistics that overlook sequential dependencies. Characterizing gaze behavior requires jointly modeling temporal fixation dynamics and the structural organization of attention across AOIs. Furthermore, eye-movement datasets in clinical settings often exhibit substantial class imbalance. This imbalance can bias model training and hinder the learning of reliable gaze representations, further complicating ASD-related detection.

To address these limitations, we propose the structural face AOI-guided Eye-Gaze Track Network (AOI-Net), a unified framework that jointly models short-term temporal dynamics and AOI-based structural organization in eye-movement sequences. By considering fixation transitions and AOI-level attentional structures as complementary views of gaze behavior, AOI-Net establishes a dual temporal-AOI modeling paradigm to learn more comprehensive representations for ASD-related gaze analysis. 
The resulting temporal and structural representations are then processed by the Temporal and Structural Experts, respectively, which refine the encoded representations and allow the model to reason over both temporal and structural aspects of gaze behavior.
A Mixture-of-Experts (MoE)~\cite{shazeer2017outrageously} mechanism then integrates these complementary representations, enabling the model to reason simultaneously over the temporal and structural aspects of gaze behavior. By combining temporal and AOI-informed structural modeling, AOI-Net provides a flexible and expressive representation of eye-movement patterns, tailored for ASD-related assessment. The main contributions are four-fold:

\begin{itemize}
    
    \item This paper introduces a unified dual temporal-AOI modeling paradigm for eye-movement representation learning, which explicitly aligns short-term fixation dynamics with AOI-level structural organization. By integrating complementary temporal and structural representations, the proposed framework enables adaptive fusion of heterogeneous gaze behaviors for ASD-related assessment.
    \item Given the discrete and fixation-driven nature of eye-movement sequences, analysis in this work reveals that prevailing long-range temporal modeling paradigms often obscure fixation-level attentional transitions that are critical for ASD discrimination. Motivated by this observation, a behavior-aware modeling strategy is introduced to explicitly capture short-range attentional dynamics.
    \item To capture ASD-related attentional organization patterns embedded in AOIs, a graph-based structural modeling strategy is introduced to construct behaviorally grounded relational connections among fixation events, enabling the model to capture clinically meaningful attention allocation and gaze interaction patterns associated with ASD.
    \item This work bridges the translational gap in AI healthcare. Unlike prior eye-movement analysis studies constrained by limited sample sizes and standardized experimental setups, our framework is validated on a large-scale and imbalanced clinical database (1,300+ participants) collected under dynamic video-based interaction paradigms, which establishes a robust methodological foundation for practical ASD screening applications.

\end{itemize}

\section{Related Work}

\subsection{Traditional Methods for ASD Diagnosis}
\label{rw}

ASD diagnosis traditionally relies on clinical observation, structured interviews, and standardized instruments such as ADOS~\cite{gotham2007autism} and ADI-R~\cite{rutter2003autism},
 which provide validated procedures for characterizing social communication deficits and restricted behaviors. These tools are considered the clinical gold standard; however, they are labor-intensive, requiring extensive professional training to administer and interpret. Diagnosis often involves prolonged sessions, multi-stage evaluations, and cross-informant reports. Such limitations constrain scalability and accessibility, particularly in regions with limited clinical resources or large screening demands.

In response to these challenges, increasing efforts have explored data-driven approaches for ASD detection using physiological and neurobiological signals as potential objective biomarkers. Neuroimaging techniques, such as functional magnetic resonance imaging (fMRI) and structural MRI (sMRI)~\cite{moridian2022automatic}, have been used to examine atypical brain connectivity and cortical development in individuals with ASD~\cite{neuroimaging, nisar2023neuroimaging}, while machine learning models applied to these data have shown promising classification performance. Similarly, electroencephalography (EEG) has been employed to analyze atypical neural oscillations and event-related potentials related to social processing~\cite{eeg,alizadehziri2026eeg}.
However, these approaches face practical limitations. Neuroimaging requires participants to remain still in confined scanning environments, which is challenging for young children and individuals with sensory sensitivities, while EEG recordings rely on electrode caps that may cause discomfort and introduce artifacts~\cite{distefano2019eeg}. These constraints motivate  exploration of alternative behavioral signals that are non-invasive, cost-effective, and ecologically valid.

\subsection{Deep Learning Approaches for Time-Series Modeling}

Time-series modeling has seen significant progress with deep learning architectures, enabling effective capture of temporal dependencies across a wide range of applications. Convolutional models such as temporal convolution networks~\cite{bai2018empirical} extract local sequential patterns efficiently, and Transformer-based frameworks~\cite{vaswani2017attention, transformerdisease} were introduced to model long-range dependencies. 
Subsequently, a variety of Transformer variants have been proposed to improve computational efficiency, strengthen long-range dependency modeling, and enrich temporal representations from different perspectives.
For example, Informer~\cite{zhou2020informer} introduces a ProbSparse self-attention mechanism together with a generative decoding strategy, enabling efficient learning of latent temporal patterns from long sequences.
Crossformer~\cite{zhang2023crossformer} further extends the Transformer by explicitly modeling cross-dimension correlations and temporal dependencies through a hierarchical cross-attention design, which enhances the representation of multivariate structures.
More recently, iTransformer~\cite{liu2023itransformer} adopts an inverted architecture tailored for sequence modeling, aiming to reduce the computational burden of vanilla Transformers while maintaining the ability to capture global temporal interactions, particularly for long horizons.
PatchTST~\cite{patchTST} segments time-series into subseries-level patches and treats them as tokens for a Transformer backbone, enabling global dependency modeling across long horizons.
In addition, XPatch~\cite{xpatch} incorporates patching into a CNN-based nonlinear stream, where the objective is not attention modeling but strengthening the extraction of local and repetitive seasonal patterns.
In the medical domain, MedSpaformer~\cite{ye2025medspaformer} adapts patch-based tokenization for clinical signals, enabling multi-granularity temporal modeling and channel-wise interaction.

However, despite the rapid advancement of general-purpose time-series models spanning from unsupervised methods for latent dynamic discovery~\cite{tan2026mask,zhang2023time} to supervised architectures for sequence prediction~\cite{xie2026anchormoe,zhang2023learning}, they are developed under assumptions that emphasize continuous signal evolution or stable long-term dependencies. Although these assumption-driven techniques have been widely and successfully applied in the medical domain~\cite{tan2025meet,xie2025de3s}, 
eye-movement data are inherently event-driven, and informative behavioral cues are often reflected in short-range temporal relationships between adjacent fixation events.
As a result, models that primarily focus on long-horizon temporal dependencies may not be well aligned with the intrinsic temporal characteristics of eye-movement behavior, where discriminative information is often localized in short temporal neighborhoods.

\subsection{Eye-Movement Data Analysis for Disease Detection}
Eye-movement data have been increasingly studied as non-invasive behavioral signals for detecting various neurological and developmental disorders, including ASD~\cite{raj2020analysis, kong2022different}, Parkinson’s disease~\cite{uribarri2023deep}, and Alzheimer’s disease~\cite{xu2024portable}.
Early studies in this area primarily relied on handcrafted gaze features~\cite{constantino2017infant, kang2020identification}. These encompass not only basic metrics like fixation duration and saccade velocity but also complex spatial-temporal descriptors such as scanpath fractal dimension and pupillary response dynamics, serving as proxies for cognitive load and autonomic nervous system regulation.
More recent approaches, such as Detach\_ROCKET~\cite{uribarri2024detach}, extend this paradigm by applying random convolutional kernels to time-series gaze data to identify the most informative features while reducing redundancy.
Most recently, deep learning approaches such as Long Short-Term Memory (LSTM) networks have been applied to eye-movement sequences to model long-range temporal dependencies and complex interactions across sequential gaze events~\cite{zhou2024gaze}. However, LSTM-based models lack explicit encoding of the discrete fixation-event and AOI structure, along with the event-driven nature of fixations and saccadic transitions that dominate eye-movement behavior.

Beyond temporal dynamics, eye-movement behavior also conveys spatial and semantic information through AOIs, which correspond to functionally meaningful regions of the visual scene. Most existing models focus on temporal dynamics and overlook the structural information encoded in AOI relationships.
Earlier studies frequently represented AOI information using aggregated descriptors, such as dwell time distributions or region-wise viewing proportions, to profile gaze preferences~\cite{wang2015atypical}. While these representations provide coarse summaries of where attention is allocated, they largely neglect the relational and transitional organization underlying gaze shifts across AOIs.
Additionally, eye-movement datasets for disease detection exhibit class imbalance, particularly in clinical settings where certain conditions are underrepresented~\cite{Santos2018Cross, Pei2021Developing}.

\begin{figure*}
  \centering
    \includegraphics[width=1\linewidth]{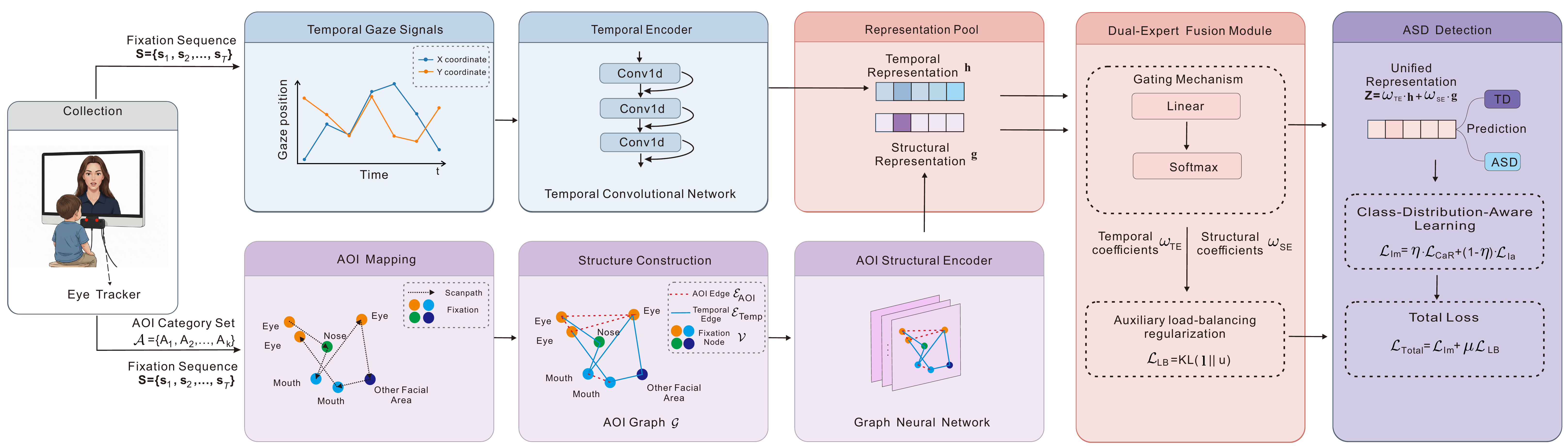}
    \caption{Overview of AOI-Net for temporal eye gaze data representation learning and ASD detection.
    }
    \label{fig:Workflow}
\end{figure*}

\section{Proposed Method}
To comprehensively characterize eye-movement behavior, we propose a structured modeling framework that jointly captures temporal dynamics and AOI-level organizational patterns. 
Short-term modeling captures sequential dependencies between consecutive fixations, reflecting moment-to-moment attentional shifts. Structural modeling incorporates AOI-level information, encoding higher-level regularities such as functional roles or task relevance.
By jointly capturing how gaze moves and where it tends to move, the framework systematically links fixation sequences with underlying AOI semantics, enabling interpretable and coherent modeling of gaze behavior. Subsequently, the two modules are dynamically integrated via a MoE mechanism, producing a unified representation where temporal sensitivity and semantic guidance are adaptively balanced.
The proposed framework preserves the semantic meaning of gaze behaviors through AOI-level structural modeling. Since each AOI corresponds to clinically meaningful facial or scene region, the constructed AOI graph can explicitly reflect how visual attention is allocated and transferred among socially relevant and non-social regions. Such relational representations enable the learned gaze patterns to be directly interpreted in terms of attention preference, fixation persistence, and transition behaviors, thereby improving the transparency and clinical interpretability of the model.

\subsection{Problem Formulation}
Eye-movement behavior exhibits a hybrid structure that combines continuous spatial-physiological measurements with discrete attentional transitions. 
Although modern eye trackers directly output fixation-level events, each fixation implicitly summarizes underlying continuous gaze dynamics and oculomotor responses. 
From a cognitive perspective, visual attention unfolds through a sequence of discrete fixation states interconnected by rapid saccadic shifts. 
Therefore, eye-movement modeling is naturally formulated as an event-driven sequential process rather than a purely continuous signal modeling task.

Let an eye-movement recording be represented as an ordered sequence of fixation events
\begin{equation}
\mathbf{S} = \{\mathbf{s}_1, \mathbf{s}_2, \ldots, \mathbf{s}_T\}, \quad \mathbf{s}_t \in \mathbb{R}^d,
\end{equation}
where $T$ denotes the total number of fixations and $\mathbf{s}_t$ is the $d$-dimensional representation of the $t$-th fixation.

Given the event-driven sequence $\mathbf{S}$, the objective is to learn a mapping function 
$f: \mathbf{S} \rightarrow y$, which predicts the clinical diagnostic label. 
Specifically, for the task of ASD screening, the formulation is defined as a binary 
classification problem
\begin{equation}
\hat{y} = f(\mathbf{S}; \theta), \quad \hat{y} \in \{0, 1\}.
\end{equation}

\subsection{Temporal Modeling of Eye-Movement}

During scene viewing, attention evolves in a rapid and incremental manner, where each fixation reflects the immediate outcome of the latest perceptual evaluation under task demands. Consequently, transitions between temporally adjacent fixations encode the most informative evidence of dynamically shifting cognitive states. In contrast, aggregating information from temporally distant events may obscure these fine-grained dynamics and introduce representational uncertainty.

Accordingly, the sequence $\mathbf{S}$ can be treated as an event-driven multivariate time series.
Eye-movement sequences are inherently event-driven, characterized by discrete fixation states with rapid saccadic transitions. This mechanism implies that information dependencies are predominantly concentrated within localized temporal neighborhoods. Specifically, transitions between adjacent or proximal fixation events reflect instantaneous shifts in attentional focus. Unlike conventional time series data, where relevance may scale with temporal span, the behavioral significance in eye-movement data is often governed by short-range, momentary dependencies.

Motivated by the event-driven nature of oculomotor behavior, we propose a new locality-constrained temporal modeling strategy tailored to eye-movement sequences. Rather than pursuing expansive temporal coverage, this stage focuses on capturing fixation-level dynamics by explicitly prioritizing local dependencies. To implement this principle, we utilize a Temporal Convolutional Network (TCN) as the temporal encoder, adapting its configuration to the localized dependency structure of oculomotor data by calibrating the growth of its receptive field to maintain a constrained dilation rate. 
Whereas standard architectures rely on large dilations to model long-range temporal context, our configuration constrains the effective receptive field to better capture short-range dependencies between successive fixation events.
This localized approach ensures that the model’s temporal representations are strictly aligned with the inherent rapid dynamics of attentional shifts.

Instead of modeling global temporal interactions across the entire sequence, we focus on learning representations from local temporal neighborhoods. The temporal encoding at the $l$-th layer is defined as:
\begin{equation}
\mathbf{h}_t^{(l)} = \sigma\!\left( \sum_{m=0}^{M-1} \mathbf{W}_m^{(l)} \, \mathbf{h}_{t - \Delta_m}^{(l-1)} + \mathbf{b}^{(l)} \right)+ \mathbf{h}_t^{(l-1)},
\end{equation}
where $\mathbf{h}_t^{(0)} = \mathbf{s}_t$, $M$ controls the width of temporal aggregation, $\{\Delta_m\}$ denotes a bounded temporal neighborhood, and $\sigma(\cdot)$ is a nonlinear activation function. In practice, this operation is implemented using a convolutional module with residual connections, forming a stack of temporal convolutional layers with a limited receptive field. This design encourages the model to capture short-term gaze transitions that are stable and behaviorally meaningful.

This temporal encoder, therefore, serves as a local dynamics extractor, providing a compact representation of short-range fixation transitions without introducing spurious long-range dependencies. Importantly, this design choice reflects a modeling stance tailored to eye-movement behavior, rather than a generic sequence modeling strategy.

\subsection{Graph Construction from Eye-Movement Sequences}

Temporal modeling captures local dynamics along the time axis; eye-movement behavior also exhibits structural dependencies that cannot be fully represented by a purely linear sequence. In particular, fixation events are not only temporally ordered but also spatially organized around semantically meaningful regions of the scene, commonly referred to as AOIs.

Let the visual scene be partitioned into a finite set of AOIs
\begin{equation}
\mathcal{A} = \{A_1, A_2, \dots, A_k\}.
\end{equation}
Each fixation at time step $t$ is associated with an AOI identity
\begin{equation}
a_t \in \mathcal{A}.
\end{equation}

Unlike continuous gaze coordinates, AOIs abstract perceptual input into semantically meaningful regions. Cognitive processes including sustained inspection, re-visitation, and region-specific attention allocation are inherently organized at this level. Therefore, AOIs reflect the structural organization of visual attention rather than merely spatial coordinates.

Eye-movement behavior often exhibits region-level coherence, including recurrent fixations within the same AOI and cyclic transitions between specific AOIs. Such patterns may span non-adjacent time steps and thus cannot be fully captured by temporal transitions alone.

A straightforward approach is to encode AOIs as categorical variables. Let the one-hot encoding of AOI identity be:
\begin{equation}
\mathbf{a}_t \in \{0,1\}^{k}.
\end{equation}

However, directly fusing categorical AOI variables with continuous fixation features inevitably introduces a representational mismatch. Numerical gaze features describe measurable behavioral attributes such as position, duration, or motion dynamics, whereas AOI identity is a symbolic indicator that conveys membership to a semantic region. When these heterogeneous data types are concatenated,
\begin{equation}
\tilde{\mathbf{s}}_t = \operatorname{concat}(\mathbf{s}_t, \mathbf{a}_t),
\end{equation}
the model is forced to process fundamentally different forms of information within a unified feature channel.

This homogenization blurs the distinction between identity-level signals and behavior-level measurements. As a result, the network may treat AOI identity as just another numerical dimension rather than as a structural cue that organizes relationships among fixation events. For example, fixation events belonging to the same AOI are not explicitly coupled; their shared regional membership does not induce interaction unless it is indirectly learned through task supervision. Consequently, region-level coherence becomes entangled with behavioral variation, leading to representational inconsistency and weakened structural modeling capacity.
To explicitly model such relationships, the eye-movement sequence is transformed into a graph representation, where fixation events are treated as nodes, and their relationships are encoded as edges. Formally, a graph is defined as:
\begin{equation}
\mathcal{G} = (\mathcal{V}, \mathcal{E}),
\end{equation}
where each node $v_t \in \mathcal{V}$ corresponds to a fixation event at time step $t$, and edges $(v_i, v_j) \in \mathcal{E}$ encode the relationships between fixation events.
This initialization assigns each node its corresponding temporal context from the original sequence, providing a temporal basis for subsequent graph modeling.

AOI-level structural relationships are first preserved through explicit edge construction. 
For two fixation nodes $v_i$ and $v_j$, an AOI-based edge is defined as:
\begin{equation}
(v_i, v_j) \in \mathcal{E}_{\text{AOI}} 
\quad \text{if} \quad a_i = a_j, \; i \neq j.
\end{equation}

These edges connect fixation events sharing the same region, regardless of temporal distance. Through such connections, region-level coherence is explicitly encoded, enabling information propagation among semantically consistent fixation events. Unlike categorical encoding, AOI-based edges introduce direct structural coupling, ensuring that shared regional membership induces interaction during representation learning.

While AOI edges preserve structural organization, eye-movement behavior remains inherently sequential. Retaining dynamics requires constructing temporal adjacency edges
\begin{equation}
(v_t, v_{t+1}) \in \mathcal{E}_{\text{Temp}}, 
\quad t = 1, \dots, T-1.
\end{equation}
These edges form the temporal backbone of the graph, preserving the ordering of fixation events and enabling localized dynamic propagation.
The final edge set is defined as:
\begin{equation}
\mathcal{E} = \mathcal{E}_{\text{AOI}} \cup \mathcal{E}_{\text{Temp}}.
\end{equation}
This unified construction jointly encodes region-level structural coherence through AOI-based edges and sequential dynamics through temporal adjacency edges. As a result, both structural and temporal dependencies are retained within a single relational framework.
To extract representations from the constructed graph, a Graph Neural Network (GNN)~\cite{scarselli2008graph} is employed, which propagates information along both temporal and AOI-based edges. At each layer, node features are updated by aggregating messages from their neighbors, allowing each fixation to integrate information from temporally adjacent fixations as well as semantically related AOI nodes. Formally, the node embedding at layer $l$ is computed as:
\begin{equation}
\mathbf{g}_i^{(l)} =\sigma\left(\mathbf{W}^{(l)}\cdot\text{AGG}\left(\left\{\mathbf{g}_j^{(l-1)} : v_j \in \mathcal{N}(v_i)\right\}\right)\right),
\end{equation}
where $\mathbf{g}_i^{(l)} \in \mathbb{R}^d$ denotes the $d$-dimensional embedding of $v_i$ at $l$-th GNN layer and $\mathcal{N}(v_i)$ denotes neighbors under AOI and temporal edges.
Through message passing, fixations integrate information from temporally adjacent events and structurally related AOI nodes. After $L$ layers, node embeddings encode fixation-level dynamics and region-level focus organization.

\begin{remark}[Cognitive and Behavioral Grounding of Graph Topology]
    The transition from a linear temporal sequence to a non-linear graph topology $\mathcal{G}$ 
    is tightly constrained by the cognitive-behavioral characteristics of ASD. Individuals with ASD often exhibit atypical gaze behaviors characterized by unstable attention maintenance, fragmented visual exploration, and recurrent shifts toward non-social or peripheral regions. 
    Classic eye-tracking studies have reported reduced fixation on socially informative regions, such as the eye region, and atypical visual scanning patterns of facial features in individuals with ASD~\cite{klin2002visual,pelphrey2002visual}. 
    These findings suggest that ASD-related gaze behaviors involve altered allocation of attention across semantically meaningful regions, with greater reliance on localized visual processing~\cite{dawson2005understanding}. 
    Such behaviors are reflected by local temporal transitions and repeated attentional allocation patterns across semantically related AOIs. 
    Accordingly, by introducing explicit AOI edges ($\mathcal{E}_{AOI}$), fixation events associated with the same semantic region are directly connected regardless of their temporal distance. This design enables the graph structure to explicitly preserve region-level attentional coherence and recurrent gaze behaviors across semantically related fixation events. Consequently, the message-passing process over $\mathcal{G}$ allows the model to jointly capture sequential fixation dynamics and higher-level AOI interaction patterns, thereby providing a behaviorally grounded representation of social attention in ASD-related gaze behavior.
\end{remark}

\subsection{Dual-Expert Fusion Module}

Sequential patterns captured by the temporal ordering of fixations and structural relations among fixations defined by AOI-based edges provide two complementary sources of information. Effective modeling of eye-movement behavior requires integrating these distinct representations. These two sources arise from distinct generative mechanisms of gaze behavior: temporal transitions emerge from rapid oculomotor control processes, whereas AOI-level coherence reflects higher-level attentional allocation and social relevance. Treating them in a single homogeneous representation risks entangling heterogeneous signals and diluting their respective inductive biases. Therefore, an integration mechanism preserving specialization while enabling adaptive coordination is required.

As a result, the MoE mechanism is employed, which enables the model to selectively leverage specialized sub-modules for different aspects of the input. Rather than statically concatenating temporal and structural embeddings, the MoE architecture allows the model to dynamically regulate the contribution of specialized sub-modules according to the characteristics of each input sequence. This design is particularly suitable for eye-movement analysis in ASD detection, where behavioral variability across individuals and stimuli is substantial. Some sequences are dominated by rapid fixation-to-fixation transitions, in which case short-term temporal dynamics provide stronger discriminative cues. In other instances, gaze behavior exhibits region-level persistence or atypical avoidance of socially salient AOIs, making structural organization more informative. A static fusion mechanism would impose uniform weighting across samples, potentially underutilizing the representational strengths of each branch. In contrast, MoE enables input-dependent expert selection, thereby increasing modeling flexibility and reducing representational interference.

Concretely, two expert modules are defined. The Temporal Expert operates on the sequential embeddings produced by the locality-constrained temporal encoder and focuses on encoding short-range dynamic dependencies between adjacent fixations. Given a fixation sequence $S$, the Temporal Expert generates a compact representation $\mathbf{h} \in \mathbb{R}^{d}$. The Structural Expert, in contrast, operates on the AOI-augmented graph representation and leverages message passing to capture cross-temporal region-level coherence. By aggregating information along both temporal and AOI-based edges, it produces a structural embedding $\mathbf{g} \in \mathbb{R}^{d}$, which encodes attentional organization beyond temporal adjacency.
To integrate these two representations, a gating mechanism is introduced to generate expert-specific weights conditioned on the input. Let $\omega_{\mathrm{TE}}$ and $\omega_{\mathrm{SE}}$ denote the routing coefficients for the temporal and structural experts, respectively. The fused representation is defined as:
\begin{equation}
\mathbf{z} = \omega_{\mathrm{TE}} \odot \mathbf{h} + \omega_{\mathrm{SE}} \odot \mathbf{g},
\end{equation}
where $\omega_{\mathrm{TE}}, \omega_{\mathrm{SE}} \in [0,1]$ are produced by a learnable gating function and $\odot$ denotes element-wise multiplication. In practice, the gating coefficients are normalized to ensure stable routing behavior. This formulation allows the model to softly allocate representational capacity between dynamic and structural cues on a per-sample basis. The gating mechanism does not merely interpolate between embeddings; it functions as a behavior-aware controller modulating the relative emphasis of short-term transitions and region-level organization according to the attentional pattern.
From a representational standpoint, the two experts encode complementary relational scales. The Temporal Expert captures transitions governed by local oculomotor dynamics, whereas the Structural Expert captures attentional organization structured by AOI semantics. By preserving this separation of inductive biases, the model avoids collapsing heterogeneous information into a single latent space prematurely. The MoE fusion stage therefore serves as a coordination layer that adaptively balances these two relational regimes, enabling the joint representation $\mathbf{z}$ to reflect instantaneous attentional shifts and structural tendencies.

\subsection{Model Training and Inference}

Clinical eye-tracking datasets for ASD assessment commonly exhibit pronounced class imbalance, where samples from the ASD group substantially outnumber those from the TD group. Such skewed distributions may bias the learned decision boundary toward the majority class and degrade the discriminative quality of the learned representations. Importantly, class imbalance affects not only the classifier layer but also the geometry of the embedding space. Minority-class samples may be compressed into poorly separated regions, leading to reduced sensitivity to subtle behavioral differences.

To address class imbalance, we employ Class-Distribution-Aware Learning (CDAL), which consists of two complementary components: 1) Class-aware Representation (CaR) and 2) Imbalance-aware (Ia).
The objective of CaR is to encourage embeddings of samples belonging to the same class to be compact while pushing apart embeddings of samples from different classes. Unlike standard cross-entropy loss, which primarily supervises the classifier logits, CaR directly constrains pairwise relationships in the feature space. This is particularly beneficial for eye-movement modeling, where subtle structural differences between ASD and TD gaze patterns must be preserved in the latent space. Formulated as:
\begin{equation}
\begin{split}
\mathcal{L}_{\text{CaR}} =& \frac{1}{n} \sum_{i=1}^{n} \left\{ \frac{1}{\alpha} \log \left[ 1 + \sum_{k \in P_i} e^{-\alpha({Sim}_{ik} - \lambda)} \right] \right. +\\ 
&\left. \frac{1}{\beta} \log \left[ 1 + \sum_{k \in N_i} e^{\beta({Sim}_{ik} - \lambda)} \right] \right\},
\end{split}
\end{equation}
where $P_i$ and $N_i$ denote positive and negative sample sets for the $i$-th sample, and ${Sim}_{ik}$ measures similarity of embeddings.

Second, the imbalance mechanism (Ia) addresses skewed distributions through inverse frequency weighting, ensuring minority classes contribute more to the loss:
\begin{equation}
    \mathcal{L}_{\text{Ia}} = -\frac{1}{N}\sum_{i=1}^{N} w_{y_i} \log p_{y_i},
\end{equation}
where $w_{y_i}$ is the class weight and $p_{y_i}$ is the predicted probability. $y_i$ denotes the ground-truth label of the $i$-th sample and $p_{y_i}$ is the predicted probability for the corresponding class. By amplifying the contribution of minority-class samples, this mechanism counteracts majority-class dominance and improves sensitivity to underrepresented behavioral patterns.

The two mechanisms are combined in a unified objective:
\begin{equation}
    \mathcal{L}_{\text{Im}} = \eta \cdot \mathcal{L}_{\text{CaR}} + (1 - \eta) \cdot \mathcal{L}_{\text{Ia}},
    \label{imloss}
\end{equation}
where $\eta \in [0,1]$ balances representation-level regularization and classification-level reweighting.
This unified objective simultaneously enforces geometric discriminability in the embedding space and statistical balance in the classifier optimization. By decoupling representation shaping from class-frequency compensation, the model becomes more robust to skewed distributions while preserving gaze-related variations relevant to ASD detection.
Beyond addressing class distribution imbalance, another challenge arises from the MoE architecture itself. Adaptive routing mechanisms in MoE architectures are known to be susceptible to expert imbalance. During training, the gating network may converge toward degenerate solutions in which one expert receives disproportionately high routing probability, suppressing the other expert and reducing specialization. Such expert dominance undermines mixture design and may lead to suboptimal generalization. To mitigate this, we introduce auxiliary load-balancing regularization.

Let $\boldsymbol{\pi} \in \mathbb{R}^{N \times E}$ denote the gating output matrix for a mini-batch over $E$ experts. Average expert load is computed as:

\begin{equation}
\mathbf{l} = \frac{1}{N} \sum_{i=1}^{N} \boldsymbol{\pi}_i.
\end{equation}

To encourage equitable utilization, we minimize the Kullback–Leibler divergence between the empirical expert load distribution $\mathbf{l}$ and a uniform prior $\mathbf{u}$, where $u_e = \frac{1}{E}$. The resulting auxiliary loss is:
\begin{equation}
\mathcal{L}_{\mathrm{LB}} = \mathrm{KL}(\mathbf{l} \,\|\, \mathbf{u}).
\end{equation}

This regularization acts as a soft constraint that penalizes persistent routing asymmetry, thereby maintaining balanced gradient flow across experts and preserving their functional specialization. Unlike hard routing constraints, this formulation encourages expert diversity instead of enforcing strict equality, while allowing data-driven adaptation. The training objective incorporates this term as a weighted component, ensuring stable optimization of the mixture framework.

The fused representation $\mathbf{z}$ is passed to task-specific classification layers for downstream ASD detection. Through the coordinated operation of temporal modeling, AOI-structured graph reasoning, and adaptive expert fusion, the proposed dual-expert framework provides a structured yet flexible representation of gaze behavior. By explicitly separating dynamic and structural inductive biases while enabling behavior-dependent integration, the model aligns closely with the hybrid nature of eye-movement generation and supports robust, interpretable ASD-related behavioral modeling. 
The final training objective combines CDAL with expert-level regularization. Specifically, the total loss is defined as:
\begin{equation}
\mathcal{L}_{\text{Total}}
= 
\mathcal{L}_{\text{Im}}
+ \mu \mathcal{L}_{\text{LB}},
\end{equation}
where $\mathcal{L}_{\text{Total}}$ denotes the overall training objective, $\mathcal{L}_{\text{Im}}$ is the imbalance-aware objective defined in Eq.~\ref{imloss}, and $\mathcal{L}_{\text{LB}}$ enforces balanced expert utilization in the MoE framework.

By jointly integrating temporal encoding, AOI-based structural modeling, and MoE fusion, the proposed framework learns discriminative and structured gaze representations while maintaining robustness to class distribution skew and preventing expert dominance, thereby enabling stable and interpretable modeling of eye-movement behaviors for ASD detection. The overall training procedure is provided in the supplementary material for completeness. The training procedure involves temporal representation learning, AOI-guided graph construction and structural representation learning, followed by adaptive fusion of temporal and structural representations through the dual-expert fusion mechanism.

\begin{table*}[!t]
\caption{ACC, F1, specificity (Spe.), and sensitivity (Sen.) comparison. \textbf{Bold}/\underline{underlined}/``-'' denote best/second-best/prediction failure.}
\label{tb:result}
\centering
\resizebox{2\columnwidth}{!}{
\begin{tabular}{l|l|cccccccc}
\toprule
Method &Metrics& Speaking & Walking1 & Walking2 & Helicopter & Baby & Tablet & Attention & Sad \\
\midrule

Informer&ACC & 0.7194 $\pm$ 0.0095 & 0.7819 $\pm$ 0.0118 & 0.7826 $\pm$ 0.0131 & 0.7515 $\pm$ 0.0084 & 0.7852 $\pm$ 0.0161 & \underline{0.8109 $\pm$ 0.0109}& 0.7345 $\pm$ 0.0085 &  0.7344 $\pm$ 0.0103 \\
~\cite{zhou2020informer} 2020&F1& 0.6657 $\pm$ 0.0089 & 0.7465 $\pm$ 0.0099 & 0.7452 $\pm$ 0.0107 & 0.7060 $\pm$ 0.0037 & 0.7603 $\pm$ 0.0070 & 0.7751 $\pm$ 0.0103 &  0.7151 $\pm$ 0.0066 & 0.7073 $\pm$ 0.0106 \\
& Spe.& 0.4941 $\pm$ 0.0194 & \underline{0.6190 $\pm$ 0.0123} & 0.5435 $\pm$ 0.0171 & 0.6190 $\pm$ 0.0195 & 0.6617 $\pm$ 0.0109 & \underline{0.6786 $\pm$ 0.0130} & 0.5366 $\pm$ 0.0159 & 0.4810 $\pm$ 0.0189 \\
& Sen. & 0.8497 $\pm$ 0.0173 & 0.8138 $\pm$ 0.0125 & 0.8272 $\pm$ 0.0111 & 0.8115 $\pm$ 0.0129 & 0.8298 $\pm$ 0.0120 & 0.8691 $\pm$ 0.0085 & 0.8032 $\pm$ 0.0141 & 0.8362 $\pm$ 0.0175 \\

\midrule
Crossformer&ACC &0.6631 $\pm$ 0.0150 &
0.6985 $\pm$ 0.0195 &
0.6944 $\pm$ 0.0186 &
0.6824 $\pm$ 0.0400 &
0.7087 $\pm$ 0.0174 &
0.7430 $\pm$ 0.0172 &
0.6457 $\pm$ 0.0057 &
0.6524 $\pm$ 0.0305 \\
~\cite{zhang2023crossformer} 2023&F1 & 0.6275 $\pm$ 0.0068 &
0.6803 $\pm$ 0.0203 &
0.6726 $\pm$ 0.0192 &
0.6607 $\pm$ 0.0322 &
0.6858 $\pm$ 0.0103 &
0.7144 $\pm$ 0.0198 &
0.6185 $\pm$ 0.0030 &
0.6245 $\pm$ 0.0253 \\
& Spe. 
& 0.5176 $\pm$ 0.0237 & 0.4286 $\pm$ 0.0411 & \textbf{0.6847} $\pm$ \textbf{0.0142} & \underline{0.7024 $\pm$ 0.0150} & 0.6951 $\pm$ 0.0202 & 0.6310 $\pm$ 0.0155 & 0.5976 $\pm$ 0.0227 & 0.5243 $\pm$ 0.0326 \\
& Sen. & 0.7254 $\pm$ 0.0195 & 0.6915 $\pm$ 0.0145 & 0.6126 $\pm$ 0.0249 & 0.6545 $\pm$ 0.0154 & 0.7500 $\pm$ 0.0181 & 0.7120 $\pm$ 0.0220 & 0.6277 $\pm$ 0.0205 & 0.7232 $\pm$ 0.0289 \\

\midrule
TimesNet&ACC & 0.7530 $\pm$ 0.0245 & 
0.7782 $\pm$ 0.0139 & 
\underline{0.7935 $\pm$ 0.0096} & 
0.7539 $\pm$ 0.0091 & 
0.7803 $\pm$ 0.0093 & 
0.8085 $\pm$ 0.0076 & 
0.7543 $\pm$ 0.0113 & 
0.7435 $\pm$ 0.0137  \\
~\cite{timesnet} 2023&F1 &0.8345 $\pm$ 0.0178 &
0.8536 $\pm$ 0.0180 &
\underline{0.8558 $\pm$ 0.0064} &
0.8366 $\pm$ 0.0087 &
\underline{0.8709 $\pm$ 0.0361}&
\underline{0.8623 $\pm$ 0.0050} &
0.8315 $\pm$ 0.0030 &
0.8205 $\pm$ 0.0108  \\
& Spe. 
& 0.5882 $\pm$ 0.0151 & 0.6071 $\pm$ 0.0151 & 0.5412 $\pm$ 0.0169 & 0.3929 $\pm$ 0.0126 & 0.5488 $\pm$ 0.0120 & 0.4762 $\pm$ 0.0146 & 0.3780 $\pm$ 0.0231 & 0.3418 $\pm$ 0.0253 \\
& Sen. & 0.7927 $\pm$ 0.0097 & 0.8564 $\pm$ 0.0101 & \textbf{0.8691 $\pm$ 0.0080} & 0.8115 $\pm$ 0.0072 & \underline{0.8777 $\pm$ 0.0050} & \underline{0.8815 $\pm$ 0.0076} & 0.8443 $\pm$ 0.0089 & 0.8136 $\pm$ 0.0112 \\
\midrule

Detach\_Rocket & ACC& \underline{0.7782 $\pm$ 0.0170} & \underline{0.7965 $\pm$ 0.0148} & 0.7814 $\pm$ 0.0117 & \underline{0.7697 $\pm$ 0.0151} & 0.7975 $\pm$ 0.0085  &  0.7928 $\pm$ 0.0063 &0.7642 $\pm$ 0.0171 & \underline{0.7591 $\pm$ 0.0060} \\
~\cite{uribarri2024detach} 2024&F1 &\underline{0.8535 $\pm$ 0.0039} & \underline{0.8585 $\pm$ 0.0120} & 0.8498 $\pm$ 0.0038 & \underline{0.8409 $\pm$ 0.0030} & 0.8567 $\pm$ 0.0025  & 0.8562 $\pm$ 0.0018 & \underline{0.8394 $\pm$ 0.0088} & \underline{0.8370 $\pm$ 0.0021} \\
& Spe. 
& 0.4471 $\pm$ 0.0343 & 0.5952 $\pm$ 0.0361 & 0.6000 $\pm$ 0.0350 & 0.4881 $\pm$ 0.0367 & 0.5610 $\pm$ 0.0317 & 0.6310 $\pm$ 0.0321 & 0.5000 $\pm$ 0.0404 & 0.4177 $\pm$ 0.0432 \\
& Sen. & \textbf{0.9171} $\pm$ \textbf{0.0277} & \textbf{0.8936} $\pm$ \textbf{0.0293} & 0.8534 $\pm$ 0.0272 & \underline{0.8586 $\pm$ 0.0263} & 0.8511 $\pm$ 0.0264 & 0.8586 $\pm$ 0.0224 & \underline{0.8777 $\pm$ 0.0310} & \underline{0.9040 $\pm$ 0.0345} \\

\midrule

EmMixformer&ACC & 0.7734 $\pm$ 0.0062 & 0.7819 $\pm$ 0.0057 & 0.7705 $\pm$ 0.0127  & 0.7685 $\pm$ 0.0021 & \underline{0.8087 $\pm$ 0.0224}& 0.7964 $\pm$ 0.0000 & \underline{0.7679 $\pm$ 0.0093} & 0.7435 $\pm$ 0.0148 \\
~\cite{qin2025emmixformer} 2025&F1 & 0.8372 $\pm$ 0.0029 & 0.8391 $\pm$ 0.0025 & 0.8243 $\pm$ 0.0109 & 0.8268 $\pm$ 0.0012 & 0.8519 $\pm$ 0.0038 & 0.8473 $\pm$ 0.0008 & 0.8303 $\pm$ 0.0044 & 0.8096 $\pm$ 0.0056 \\
& Spe. 
& 0.5176 $\pm$ 0.0461 & \underline{0.6190 $\pm$ 0.0342} & 0.6706 $\pm$ 0.0493 & 0.6310 $\pm$ 0.0419 & 0.6707 $\pm$ 0.0331 & 0.6190 $\pm$ 0.0410 & 0.5854 $\pm$ 0.0477 & \underline{0.5443 $\pm$ 0.0442} \\
& Sen. & \underline{0.9016 $\pm$ 0.0427} & 0.8564 $\pm$ 0.0264 & 0.8063 $\pm$ 0.0441 & 0.8482 $\pm$ 0.0420 & 0.8032 $\pm$ 0.0181 & 0.8691 $\pm$ 0.0291 & 0.8564 $\pm$ 0.0453 & 0.8531 $\pm$ 0.0366 \\

\midrule
MPTSNet&ACC & 0.6199 $\pm$ 0.0270 & 0.6458 $\pm$ 0.0368 & 0.6848 $\pm$ 0.0166 & 0.6497 $\pm$ 0.0588 & 0.6407 $\pm$ 0.0162 & 0.6860 $\pm$ 0.0147 & 0.6568 $\pm$ 0.0114 & 0.5899 $\pm$ 0.0334 \\
~\cite{mu2025mptsnet} 2025&F1 & 0.6949 $\pm$ 0.0163 & 0.6964 $\pm$ 0.0231 & 0.7515 $\pm$ 0.0160 & 0.7176 $\pm$ 0.0354 & 0.6816 $\pm$ 0.0076 & 0.7695 $\pm$ 0.0165 & 0.7364 $\pm$ 0.0223 & 0.6723 $\pm$ 0.0548 \\
& Spe.
& \textbf{0.7765} $\pm$ \textbf{0.0361} & \underline{0.6190 $\pm$ 0.0189} & \underline{0.6824 $\pm$ 0.0281} & 0.6667 $\pm$ 0.0091 & \underline{0.7561 $\pm$ 0.0046} & 0.6429 $\pm$ 0.0162 & \underline{0.6341 $\pm$ 0.0272} & \textbf{0.5570} $\pm$ \textbf{0.0245} \\
& Sen. & 0.3990 $\pm$ 0.0375 & 0.6649 $\pm$ 0.0436 & 0.6545 $\pm$ 0.0249 & 0.7435 $\pm$ 0.0077 & 0.7074 $\pm$ 0.0082 & 0.6859 $\pm$ 0.0173 & 0.6809 $\pm$ 0.0211 & 0.6610 $\pm$ 0.0194 \\
\midrule

XPatch&ACC &0.6966 $\pm$ 0.0090 & 0.6752 $\pm$ 0.0056 & 0.7343 $\pm$ 0.0076 & 0.7006 $\pm$ 0.0084 & 0.6852 $\pm$ 0.0098 & 0.7030 $\pm$ 0.0151 & 0.7025 $\pm$ 0.0093 & 0.6523 $\pm$ 0.0104 \\
~\cite{xpatch} 2025&F1 & 0.7732 $\pm$ 0.0053 & 0.7008 $\pm$ 0.0171 & 0.7882 $\pm$ 0.0040 & 0.7562 $\pm$ 0.0075 & 0.7240 $\pm$ 0.0061 & 0.7575 $\pm$ 0.0179 & 0.7670 $\pm$ 0.0111 & 0.7022 $\pm$ 0.0125 \\
& Spe.
& 0.3412 $\pm$ 0.0308 & 0.5238 $\pm$ 0.0209 & 0.5059 $\pm$ 0.0178 & 0.4524 $\pm$ 0.0195 & 0.5854 $\pm$ 0.0238 & 0.4881 $\pm$ 0.0167 & 0.4146 $\pm$ 0.0226 & 0.4810 $\pm$ 0.0329 \\
& Sen. & 0.8653 $\pm$ 0.0300 & 0.8191 $\pm$ 0.0184 & 0.8586 $\pm$ 0.0156 & 0.8396 $\pm$ 0.0209 & 0.7394 $\pm$ 0.0241 & 0.8143 $\pm$ 0.0153 & 0.8457 $\pm$ 0.0214 & 0.8192 $\pm$ 0.0255 \\

\midrule
MedSpaformer&ACC & 0.6942 $\pm$ 0.0000 & 0.6716 $\pm$ 0.0057 & 0.6920 $\pm$ 0.0000 & 0.6945 $\pm$ 0.0000 & 0.6963 $\pm$ 0.0000 & 0.6982 $\pm$ 0.0297 & 0.6963 $\pm$ 0.0000 & 0.6862 $\pm$ 0.0090 \\
~\cite{ye2025medspaformer} 2026&F1 & 0.4098 $\pm$ 0.0000 & 0.5830 $\pm$ 0.0029 & 0.4090 $\pm$ 0.0000& 0.4099 $\pm$ 0.0000 & 0.4105 $\pm$ 0.0000 & 0.6250 $\pm$ 0.0308 & 0.4105 $\pm$ 0.0000 & 0.4705 $\pm$ 0.0942\\
& Spe. 
 & $-$ & 0.3452 $\pm$ 0.0458 & $-$ & $-$ & $-$ & 0.4762 $\pm$ 0.0247 & $-$ & 0.3671 $\pm$ 0.0447 \\
& Sen.& $-$ & 0.8085 $\pm$ 0.0411 & $-$ & $-$ & $-$ & 0.7487 $\pm$ 0.0205 & $-$ & 0.8136 $\pm$ 0.0394 \\

\midrule
AOI-Net  & ACC&\textbf{0.8201 $\pm$ 0.0102} & \textbf{0.8248 $\pm$ 0.0017} & \textbf{0.8140 $\pm$ 0.0017} & \textbf{0.8145 $\pm$ 0.0030} & \textbf{0.8358 $\pm$ 0.0017} & \textbf{0.8364 $\pm$ 0.0030} & \textbf{0.7938 $\pm$ 0.0046} & \textbf{0.7904 $\pm$ 0.0080} \\
(ours)&F1 & \textbf{0.8637 $\pm$ 0.0045} & \textbf{0.8739 $\pm$ 0.0042} & \textbf{0.8714 $\pm$ 0.0018} & \textbf{0.8647 $\pm$ 0.0030} & \textbf{0.8835 $\pm$ 0.0036} & \textbf{0.8813 $\pm$ 0.0008} & \textbf{0.8551 $\pm$ 0.0087} & \textbf{0.8561 $\pm$ 0.0051} \\
& Spe.
& \underline{0.6353 $\pm$ 0.0192} 
& \textbf{0.7024 $\pm$ 0.0541} 
& 0.5961 $\pm$ 0.0563 
& \textbf{0.7103 $\pm$ 0.0112} 
& \textbf{0.7602 $\pm$ 0.0058} 
& \textbf{0.7183 $\pm$ 0.0112} 
& \textbf{0.6554 $\pm$ 0.0392} 
& 0.5232 $\pm$ 0.0158 \\
& Sen.
& \underline{0.9016 $\pm$ 0.0085} 
& \underline{0.8794 $\pm$ 0.0262} 
& \underline{0.8687 $\pm$ 0.0025} 
& \textbf{0.8659 $\pm$ 0.0136} 
& \textbf{0.9096 $\pm$ 0.0075} 
& \textbf{0.8866 $\pm$ 0.0065} 
& \textbf{0.8898 $\pm$ 0.0176} 
& \textbf{0.9096 $\pm$ 0.0148} \\

\bottomrule
\end{tabular}
}
\end{table*}

\begin{table}[!t]
\caption{Comparison of ablated AOI-Net variants. \textbf{Bold}/\underline{underlined} denote best/second-best results, respectively.}
\label{tb:ab_acc}
\centering
\scriptsize

\begin{tabular}{l |l| c c c}
\toprule
\multirow{2}{*}{Dataset}&\multirow{2}{*}{Metric} & \multicolumn{3}{c}{Variants} \\
\cmidrule(lr){3-5}
&& AOI-Net & Variant I & Variant II \\
\midrule
\multirow{2}{*}{Speaking} & ACC & \textbf{0.8201} & \underline{0.7950} & 0.7950 \\
 & F1 & \textbf{0.8637} & 0.8571 & \underline{0.8589} \\
\midrule
\multirow{2}{*}{Walking1} & ACC & \textbf{0.8248} & \underline{0.8076} & 0.8064 \\
 & F1 & \textbf{0.8739} & \underline{0.8641} & 0.8594 \\
\midrule
\multirow{2}{*}{Walking2} & ACC & \textbf{0.8140} & \underline{0.8128} & 0.8056 \\
 & F1 & \textbf{0.8714} & \underline{0.8688} & 0.8655 \\
\midrule
\multirow{2}{*}{Helicopter} & ACC & \textbf{0.8145} & \underline{0.8012} & 0.7988 \\
 & F1 & \textbf{0.8647} & \underline{0.8595} & 0.8591 \\
\midrule
\multirow{2}{*}{Baby} & ACC & \textbf{0.8358} & \underline{0.8346} & 0.8333 \\
 & F1 & \textbf{0.8835} & \underline{0.8813} & 0.8785 \\
\midrule
\multirow{2}{*}{Tablet} & ACC & \textbf{0.8364} & 0.8364 & 0.8279 \\
 & F1 & \underline{0.8813} & \textbf{0.8835} & 0.8732 \\
\midrule
\multirow{2}{*}{Attention} & ACC & \textbf{0.7938} & \underline{0.7914} & \underline{0.7914} \\
 & F1 & \textbf{0.8551} & \underline{0.8524} & 0.8500 \\
\midrule
\multirow{2}{*}{Sad} & ACC & \underline{0.7904} & \textbf{0.7943} & 0.7878 \\
 & F1 & \textbf{0.8561} & \underline{0.8518} & 0.8494 \\
\midrule
\multirow{2}{*}{Avg. Rank} & ACC & 1.13 & 1.75 & 2.75 \\
 & F1 & 1.13 & 1.88 & 2.88 \\
\bottomrule
\end{tabular}
\end{table}

\section{Experiment}

To systematically evaluate AOI-Net, we consider seven aspects: comparative ASD detection, AOI importance and prototype behavior, ablation, GNN propagation depth, efficiency, temporal modeling scale, and generalization analysis under dataset exclusion. \href{https://github.com/Zhanpei-ai/CIM-AOI-Net/tree/main/Supplement}{Supplementary Materials} provide the complete training algorithm, detailed participant statistics, sensitivity analyses of $\eta$ and AOI importance, and ablations of fusion strategies and gate activation functions.

\subsection{Experimental Setup}
\textbf{Model Configuration:} Eye-movement data are analyzed via a temporal expert and a structural expert, and their learned representations are integrated by AOI-Net, incorporating Class-Distribution-Aware Learning. $\eta$ for controlling the contribution of CaR and Ia is set to 0.25, and $\mu$ is set to 0.01 across all datasets. Adam optimizer is used to optimize the model.

\textbf{Counterparts:} Informer~\cite{zhou2020informer}, Crossformer~\cite{zhang2023crossformer}, TimesNet~\cite{timesnet}, MPTSNet~\cite{mu2025mptsnet}, and XPatch~\cite{xpatch} are general time-series analysis models. MedSpaformer~\cite{ye2025medspaformer} is designed for medical time-series classification. For eye-movement data analysis, EmMixformer~\cite{qin2025emmixformer} is proposed for biometric identification, while Detach\_Rocket~\cite{uribarri2024detach} has been applied to Parkinson's disease recognition. F1-score~\cite{f1}, Accuracy (ACC)~\cite{acc}, Sensitivity (Sen.), and Specificity (Spe.) are adopted as evaluation metrics. All experiments are conducted on a machine equipped with an NVIDIA RTX 4090 GPU.

\textbf{Dataset:}
We used eight eye-movement datasets, namely Speaking, Walking1, Walking2, Helicopter, Baby, Tablet, Attention, and Sad, collected by the Shenzhen Maternity and Child Healthcare Hospital. These datasets correspond to different experimental paradigms capturing diverse eye-movement patterns under dynamic visual stimuli. ASD participants were clinically diagnosed by licensed clinicians, while TD participants were screened to exclude known neurodevelopmental disorders. 
The processed fixation sequences were mapped to predefined AOIs, which were consistently annotated across participants and used as input representation for the proposed model.
The objective of these datasets is to support ASD-versus-TD identification based on eye-movement patterns.

\subsection{Comparative Study of ASD Detection}

The detection performance of different algorithms is evaluated using ACC, F1-score, Sen., and Spe. to provide a comprehensive assessment of clinical screening performance. In addition, Receiver Operating Characteristic (ROC) curves are plotted to further evaluate the discrimination capability of different methods under varying decision thresholds. Due to space limitations, representative ROC results on two benchmark datasets are reported.

Overall Performance: As presented in Table~\ref{tb:result}, the proposed AOI-Net consistently achieves superior detection performance across all eight datasets in terms of both ACC and F1-score. Quantitatively, AOI-Net surpasses the second-best performing methods by substantial margins in the majority of experimental scenarios. The result demonstrates the capability to handle the heterogeneity of eye-movement patterns elicited by different paradigms, demonstrating a robust generalization capability that is critical for clinical screening applications.

Model comparison: A comparison with baseline methods highlights the limitations of general-purpose time-series models in this task. Transformer-based architectures (e.g., Informer and Crossformer) underperform compared to AOI-Net, likely due to the event-driven and short-range nature of gaze behavior. By emphasizing global temporal dependencies, these models may overestimate weak long-range correlations and introduce irrelevant information. Detach\_Rocket, which captures local temporal patterns through random convolutional kernels, achieves relatively better results but still lags behind AOI-Net, indicating that local features alone are insufficient. The superior performance of AOI-Net stems from its ability to jointly model local temporal dynamics and the structural organization of AOIs, yielding a more discriminative representation of autistic gaze patterns.

\begin{figure}

    \centering
    \includegraphics[width=1\linewidth]{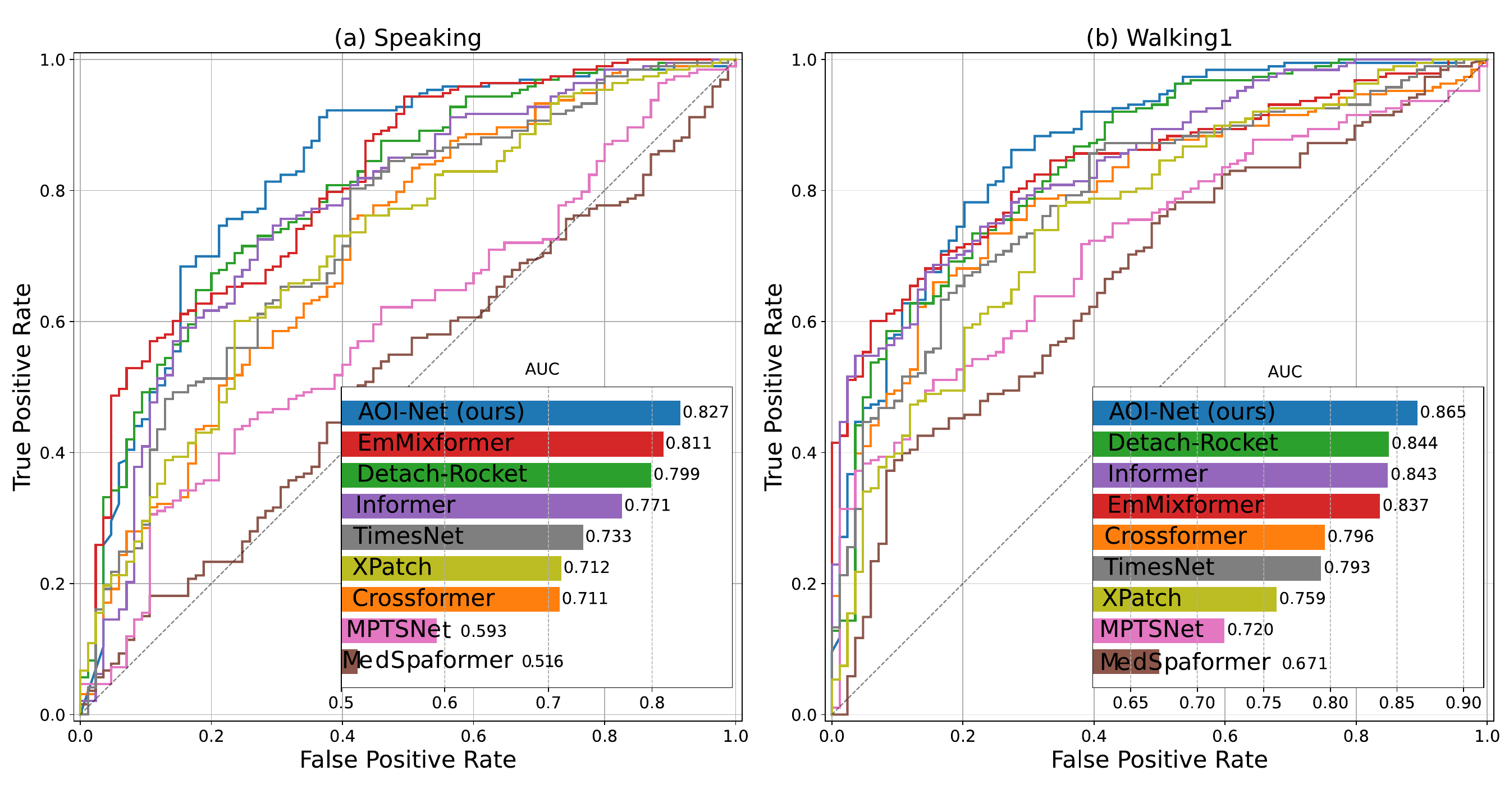}
    \caption{ROC curves and AUC of different methods on Speaking and Walking1. The histogram in the figure also serves as the legend for the ROC curves.}
    \label{fig:roc}

\end{figure}

Screening Reliability: From a clinical screening perspective, Sen. is the most critical metric, as ASD screening systems are primarily designed to minimize missed diagnoses and ensure that potential ASD cases are correctly identified for further clinical evaluation. As shown in Table~\ref{tb:result}, AOI-Net consistently achieves the best or near-best Sensitivity across most datasets, demonstrating its strong capability in capturing subtle and discriminative atypical gaze patterns associated with ASD.
In contrast, several competing methods that achieve relatively higher Spe. often exhibit noticeably lower Sensitivity, indicating a tendency to correctly reject negative cases at the cost of missing true ASD cases. Such behavior is suboptimal for screening, where false negatives are clinically more costly than false positives, as missed cases delay early intervention and diagnosis.
Overall, AOI-Net prioritizes high Sensitivity while maintaining competitive Specificity, leading to a more appropriate trade-off for real-world ASD screening applications where early detection is the primary objective.

ROC Analysis: To examine the screening capability of different methods across varying decision thresholds, ROC curves on two datasets are illustrated in Fig.~\ref{fig:roc}. It can be observed that the ROC curves of AOI-Net consistently dominate those of competing approaches over most operating regions, indicating a superior balance between Sensitivity and Specificity. Moreover, AOI-Net achieves the largest AUC values, demonstrating stronger discriminative ability between ASD and TD. The improvement in both ROC shape and AUC validates that the proposed AOI-guided temporal-structural representation effectively captures ASD-related gaze characteristics and remains robust under different screening thresholds.

\begin{figure}

    \centering
    \includegraphics[width=1\linewidth]{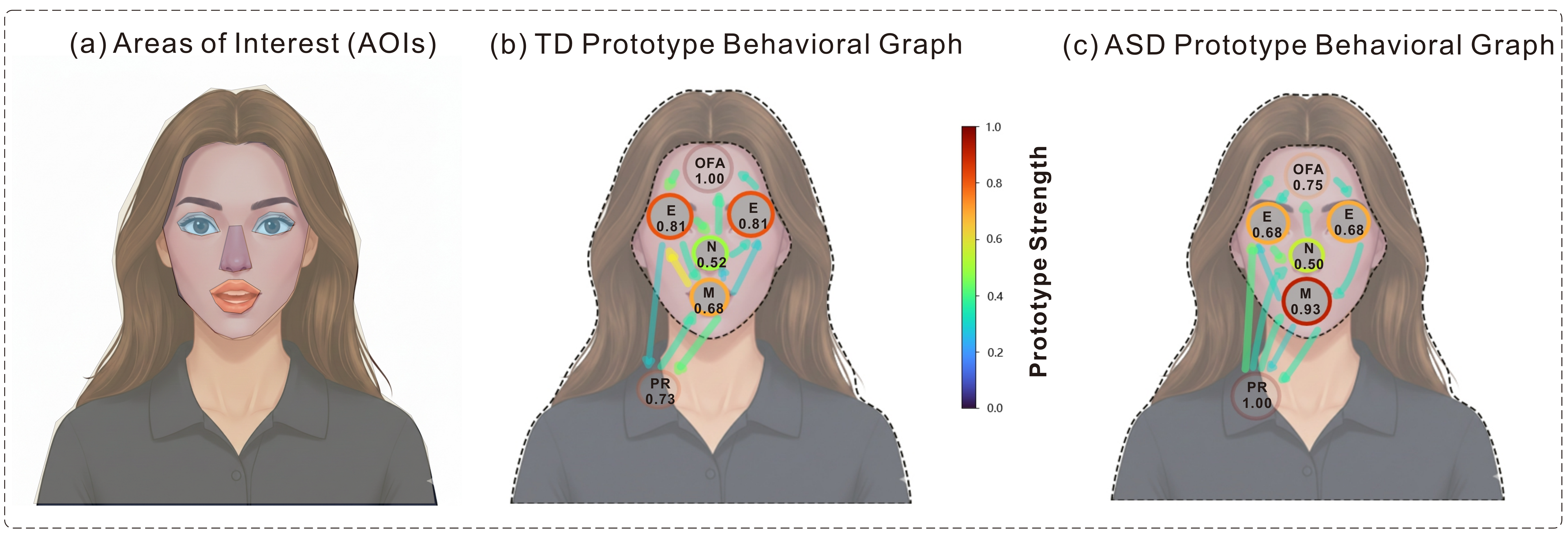}
    \caption{Behavioral graphs for TD (b) and ASD (c) participants on the Speaking dataset. Nodes: E (Eye), N (Nose), M (Mouth), OFA (Other Facial Area), PR (Peripheral Region); node size encodes attention strength (values shown). Edge color (cool-warm) indicates transition strength; self‑loops represent staying in the same region. Edges with weight $ <  0.05$ are omitted.}
    \label{fig:prototype}

\end{figure}
\subsection{Analysis of AOI Importance and Prototype Behavior}

To further investigate the learned gaze behavior patterns, we additionally visualize the prototype behavioral graphs learned by the proposed framework in Fig.~\ref{fig:prototype}. Visualization reveals that TD participants exhibit denser and more balanced gaze transitions across multiple socially salient facial regions, especially between the two eyes and other central facial areas. In contrast, ASD participants demonstrate relatively restricted facial exploration behaviors, characterized by weaker transition connectivity among socially relevant AOIs and stronger concentration toward peripheral or non-social regions.
Moreover, compared with the TD group, ASD samples exhibit reduced transition diversity and weaker global facial interaction patterns. Even when facial regions are attended, the gaze behaviors of ASD participants tend to remain localized rather than dynamically coordinated across multiple informative facial regions. Such behavioral characteristics are consistent with established findings in autism-related studies, where individuals with ASD often demonstrate reduced eye-contact preference, atypical social attention allocation, and increased focus on repetitive or non-social visual stimuli.
Therefore, both the AOI importance distributions and the prototype behavioral graph visualizations provide clinically meaningful evidence that the proposed AOI-Net framework captures interpretable and behaviorally relevant gaze interaction patterns associated with ASD.

\begin{table}[!t]
\caption{AOI edge strategies: KNN, Dynamic, Weighted, and Full. \textbf{Bold}/\underline{underlined} denote best/second-best.}
\label{tb:ab_edge}
\centering
\setlength{\tabcolsep}{3pt}
\renewcommand{\arraystretch}{1.05}
\resizebox{1.0\columnwidth}{!}{
\begin{tabular}{c|c|ccc|ccc}
\toprule
\multirow{2}{*}{Dataset}
& \multirow{2}{*}{Metric}
& \multicolumn{3}{c|}{\shortstack{KNN-Based \\Construction Strategy}}
& \multicolumn{3}{c}{\shortstack{AOI Edge \\Construction Strategy}} \\
\cmidrule(lr){3-5}
\cmidrule(lr){6-8}
& &
\makebox[1.1cm][c]{\shortstack{KNN\\($k=2$)}}
& \makebox[1.1cm][c]{\shortstack{KNN\\($k=4$)}}
& \makebox[1.1cm][c]{\shortstack{KNN\\($k=6$)}}
& \makebox[1.25cm][c]{\shortstack{AOI-Net\\(Dynamic)}}
& \makebox[1.25cm][c]{\shortstack{AOI-Net\\(Weighted)}}
& \makebox[1.25cm][c]{\shortstack{AOI-Net\\(Full)}} \\
\midrule

\multirow{2}{*}{Speaking}
& ACC
& 0.7482
& 0.7662
& 0.7482
& 0.8034
& \underline{0.8165}
& \textbf{0.8201} \\
& F1
& 0.8293
& 0.8441
& 0.8293
& \underline{0.8646}
& \textbf{0.8741}
& 0.8637 \\

\midrule

\multirow{2}{*}{Walking1}
& ACC
& 0.7941
& 0.7978
& 0.7904
& \underline{0.8199}
& 0.8113
& \textbf{0.8248} \\
& F1
& 0.8470
& 0.8533
& 0.8480
& \underline{0.8655}
& 0.8620
& \textbf{0.8739} \\

\midrule

\multirow{2}{*}{Walking2}
& ACC
& 0.7971
& 0.8007
& 0.7971
& \underline{0.8056}
& 0.8007
& \textbf{0.8140} \\
& F1
& 0.8634
& 0.8655
& 0.8586
& \underline{0.8669}
& 0.8595
& \textbf{0.8714} \\

\midrule

\multirow{2}{*}{Helicopter}
& ACC
& \textbf{0.8182}
& 0.8000
& 0.8000
& 0.8000
& 0.8109
& \underline{0.8145} \\
& F1
& \underline{0.8656}
& 0.8533
& 0.8541
& 0.8550
& \textbf{0.8657}
& 0.8647 \\

\midrule

\multirow{2}{*}{Baby}
& ACC
& 0.8296
& 0.8222
& 0.8111
& 0.8247
& \underline{0.8321}
& \textbf{0.8358} \\
& F1
& 0.8810
& 0.8710
& 0.8595
& 0.8719
& \underline{0.8813}
& \textbf{0.8835} \\

\midrule

\multirow{2}{*}{Tablet}
& ACC
& 0.8255
& \underline{0.8327}
& 0.8073
& 0.8303
& 0.8230
& \textbf{0.8364} \\
& F1
& 0.8750
& 0.8736
& 0.8556
& \underline{0.8777}
& 0.8694
& \textbf{0.8813} \\

\midrule

\multirow{2}{*}{Attention}
& ACC
& 0.7741
& 0.7815
& 0.7667
& \textbf{0.8000}
& 0.7914
& \underline{0.7938} \\
& F1
& 0.8448
& 0.8451
& 0.8205
& \textbf{0.8631}
& 0.8537
& \underline{0.8551} \\

\midrule

\multirow{2}{*}{Sad}
& ACC
& 0.7539
& 0.7539
& 0.7578
& \underline{0.7747}
& \underline{0.7747}
& \textbf{0.7904} \\
& F1
& 0.8372
& 0.8293
& 0.8278
& 0.8378
& \underline{0.8447}
& \textbf{0.8561} \\

\midrule

\multirow{2}{*}{Average Rank}
& ACC
& 4.31
& 4.13
& 5.50
& 2.81
& 3.00
& 1.25 \\
& F1
& 4.06
& 4.38
& 5.69
& 2.50
& 2.75
& 1.63 \\

\bottomrule
\end{tabular}
}
\end{table}

\subsection{Ablation Study}

\textbf{Effectiveness of AOI and Temporal Edges:}
To quantitatively assess the contributions of individual graph components, ablation studies were conducted on two distinct variants: Variant ${I}$, which removes the AOI-based edges (retaining only temporal connections); and Variant ${II}$, which further removes the temporal edges from Variant ${I}$. The results are summarized in Table~\ref{tb:ab_acc}.
Since both Variant ${I}$ and ${II}$ exclude AOI edges and thus lack explicit AOI relational information, the AOI features are directly provided to the temporal modeling module for processing.
The ablation results provide deeper insights into the specific contribution of each modeling dimension:

First, the contribution of the AOI-based structural graph is analyzed. Variant ${I}$, which models fixations solely based on temporal adjacency without explicit AOI structural links, exhibits a discernible performance degradation compared to the full AOI-Net. 
This decline corroborates the hypothesis that explicit modeling of AOI relationships is crucial for ASD detection. AOI edges enable the network to encode high-level scanpath strategies—specifically, how subjects transition between semantically significant regions. 

Second, the effect of temporal graph modeling is further examined by Variant ${II}$, which removes the temporal edges on top of Variant ${I}$, effectively eliminating the Structural Expert and discarding graph-based relational modeling altogether.
Compared to both the full AOI-Net and Variant ${I}$, Variant ${II}$ exhibits a more pronounced performance degradation. This result indicates that the structural expert plays a critical role in modeling relational dependencies among fixations. Without temporal edges and graph message passing, the model reduces to a purely sequential architecture, lacking the ability to explicitly encode fixation-to-fixation interactions.

The inferior performance of Variant ${II}$ demonstrates that simply providing AOI features to a temporal module is insufficient. Explicit relational modeling through the GNN expert enhances representation learning by capturing structured dependencies that cannot be fully recovered through sequential modeling alone. Their integration enables AOI-Net to effectively capture the joint spatial-semantic and relational dynamics underlying autistic gaze behavior.

\textbf{Analysis of AOI Edge Construction Strategy:}

To further examine the contribution of AOI semantic connections, we compared AOI-Net with three alternative edge construction strategies: KNN-based connections, dynamic AOI edges, and weighted AOI edges. For KNN-based variants, different neighborhood sizes ($k=2,4,6$) were investigated, where each fixation node selects neighboring fixation points based on spatial proximity without considering AOI semantic correspondence. As $k$ increases, the graph gradually incorporates more neighborhood information. For dynamic edge variants, the model starts from candidate AOI connections and adaptively determines the existence of each connection during training according to the observed gaze patterns.  For the weighted edge variant, the original AOI connections are preserved, while learnable weights are assigned to different AOI relationships to adjust their contributions during message propagation. The weights are optimized jointly with the model parameters based on the characteristics of connected fixation nodes, allowing the model to further optimize the strength of semantic connections. The results are summarized in Table~\ref{tb:ab_edge}.
The results show that KNN-based graphs achieve reasonable performance, indicating that generic neighborhood information can provide useful relational cues. However, as $k$ increases, connections are gradually expanded based on spatial proximity without considering AOI semantic correspondence. Such additional neighbors may include visually or behaviorally less relevant fixation points, reducing the discriminative capability of message aggregation. The dynamic edge variant provides greater flexibility by learning input-dependent connections, but its performance remains inferior, suggesting that purely data-driven edge adaptation may reduce the consistency of AOI-level semantic relationships. Similarly, the weighted edge variant preserves the original AOI connections while introducing learnable weights to adjust the contribution of different AOI relationships during message propagation, but does not outperform the predefined AOI edges. However, since AOI annotations already provide reliable semantic priors by capturing region-level attentional correspondence, additional flexibility in edge weighting may emphasize sample-specific variations and limit the improvement over the predefined AOI structure. In contrast, the predefined AOI edge formulation preserves semantic organization among attentional regions and provides stable structural guidance for ASD-related gaze modeling.

 \begin{figure}
    \centering
    \includegraphics[width=1\linewidth]{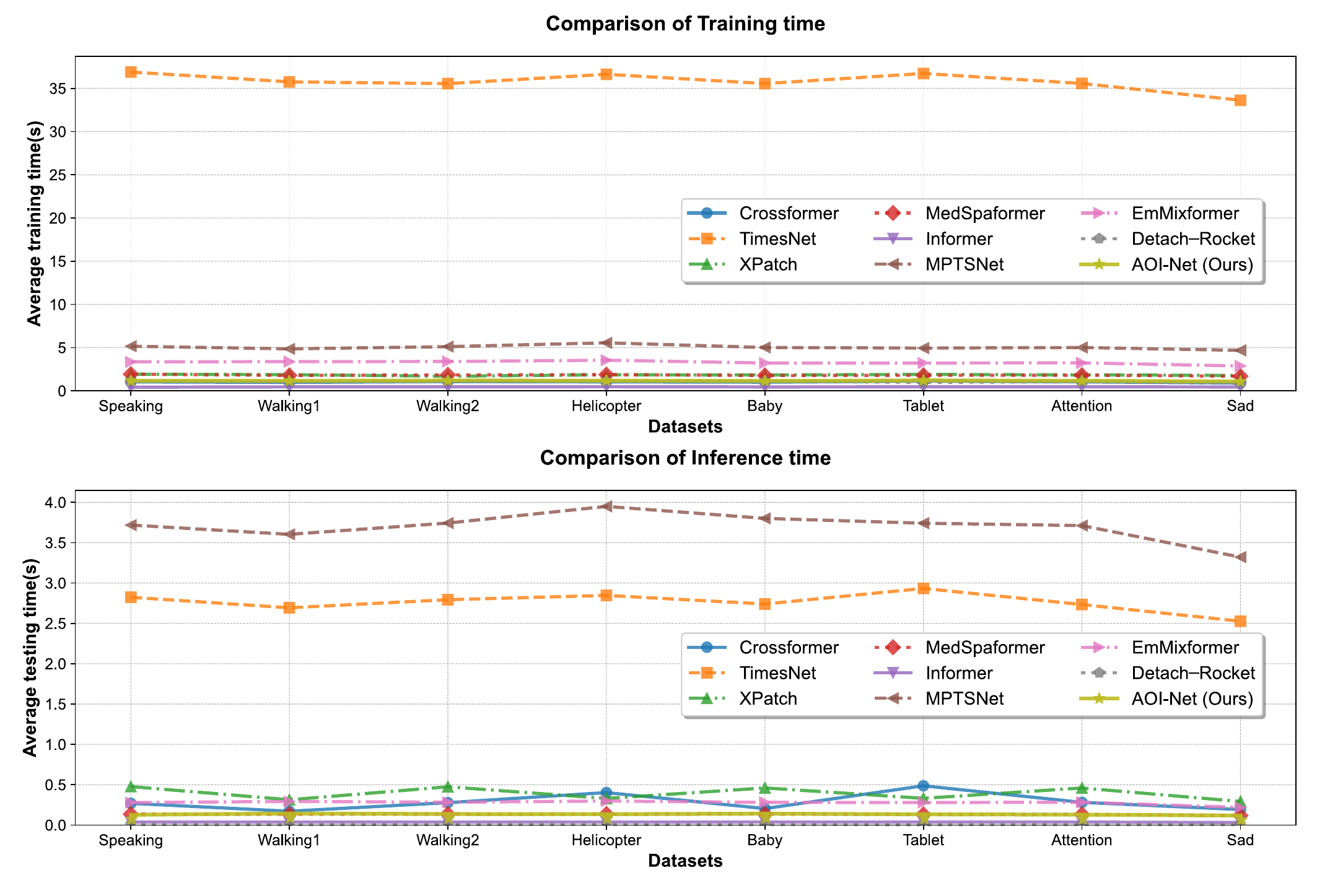}
    \caption{Efficiency comparison of training and inference time.}
    \label{fig:efficiency}
\end{figure}

\subsection{Impact of GNN Propagation Depth}
\begin{figure}

    \centering
    \includegraphics[width=1\linewidth]{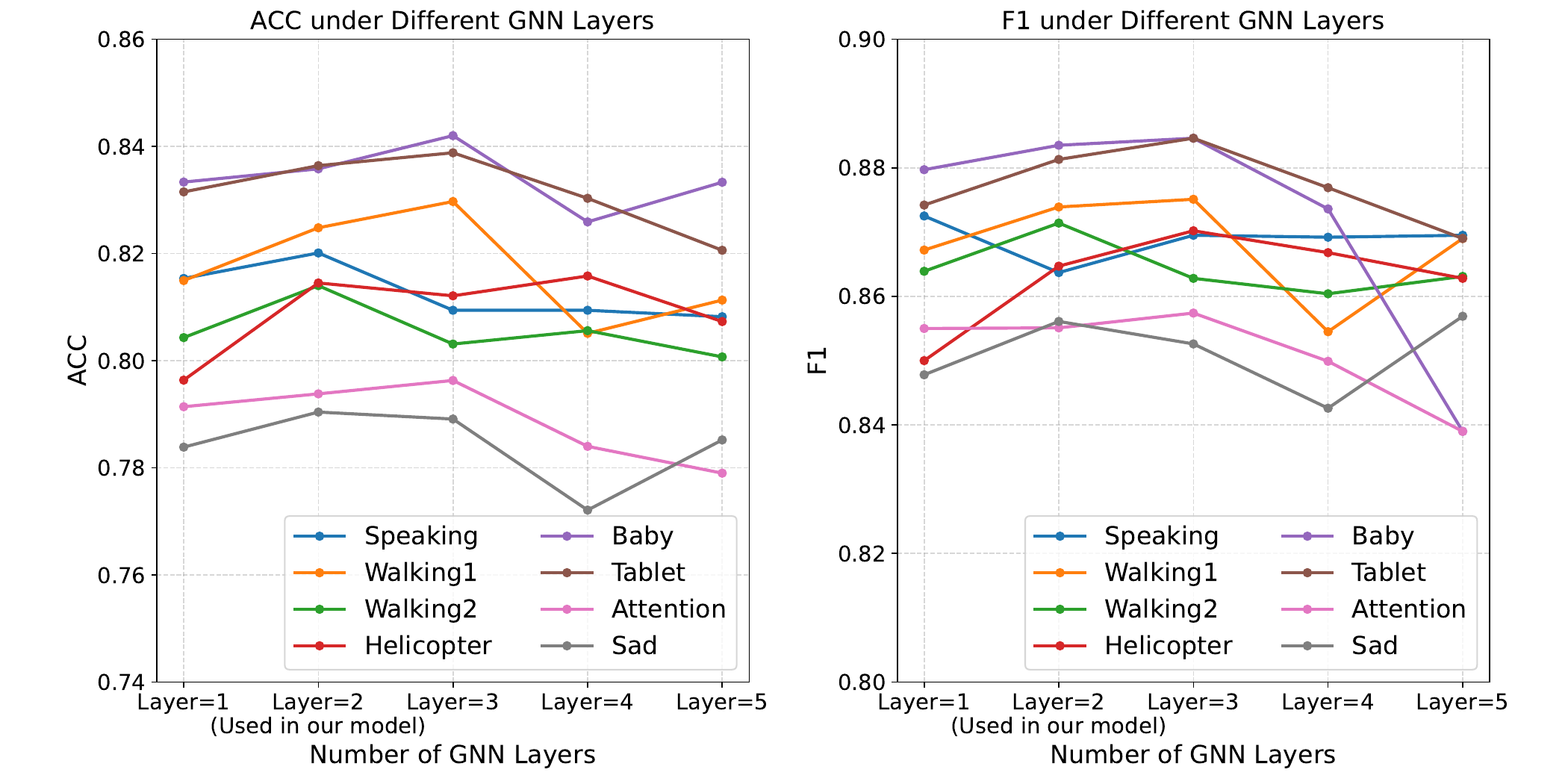}
    \caption{Comparison of different numbers of GNN layers across datasets. }
    \label{fig:gnn}

\end{figure}

To further investigate the structural modeling behavior of the proposed AOI-Net, we additionally analyze the impact of GNN propagation depth by varying the number of graph layers from 1 to 5. The results are summarized in Fig.~\ref{fig:gnn}.

The experimental results show a clear trend where detection performance first improves and then gradually declines as the number of GNN layers increases. Shallow graph propagation significantly enhances the representation capability compared with a single-layer configuration, indicating that moderate structural aggregation is beneficial for modeling AOI-level attentional relationships. Allowing fixation nodes to interact with semantically related AOIs and neighboring fixation events, the structural encoder can effectively capture region-level attentional organization patterns associated with ASD.

However, when the graph depth becomes excessively large, the performance gradually decreases. This observation suggests that excessively expanding the propagation range may dilute discriminative local attentional patterns that are behaviorally meaningful for ASD screening. 
These findings support the design choice of using controlled graph propagation in AOI-Net, and indicate that shallow propagation is sufficient to aggregate behaviorally meaningful information while avoiding unnecessary feature mixing caused by excessive message passing.
These findings further validate that the proposed AOI-guided graph construction is not merely a generic graph modeling strategy, but a behavior-aware structural representation mechanism specifically designed for ASD-related gaze analysis. The optimal performance achieved with relatively shallow graph layers also aligns with cognitive observations that ASD-related gaze behaviors are characterized by localized and fragmented attentional organization patterns rather than globally coordinated long-range interactions.

\subsection{Efficiency Analysis}
The efficiency comparison results are illustrated in Fig.~\ref{fig:efficiency}, which reports the average training time and testing time across different datasets. From the training efficiency perspective, TimesNet and MPTSNet exhibit significantly higher training costs compared with other methods, indicating heavier model complexity and computational overhead. In contrast, most transformer-based baselines such as Crossformer, XPatch, MedSpaformer, Informer, and EmMixformer maintain relatively low and stable training times across datasets. The proposed AOI-Net demonstrates competitive training efficiency, remaining consistently close to the lower range of the compared methods while avoiding the large computational burden of deep architectures. This suggests that the dual-expert design does not introduce substantial additional training cost.

A similar trend can be observed in the inference phase. Methods such as MPTSNet and TimesNet require noticeably longer inference time, whereas the majority of models show relatively efficient testing performance. AOI-Net achieves stable and low inference latency across all datasets, remaining comparable to or better than several existing methods. This indicates that the proposed framework maintains efficient inference while incorporating both temporal dynamics and AOI-level structural modeling. Overall, the results demonstrate that AOI-Net achieves a favorable balance between computational efficiency and modeling capability, making it suitable for practical ASD detection scenarios.

\begin{figure}

    \centering
    \includegraphics[width=1\linewidth]{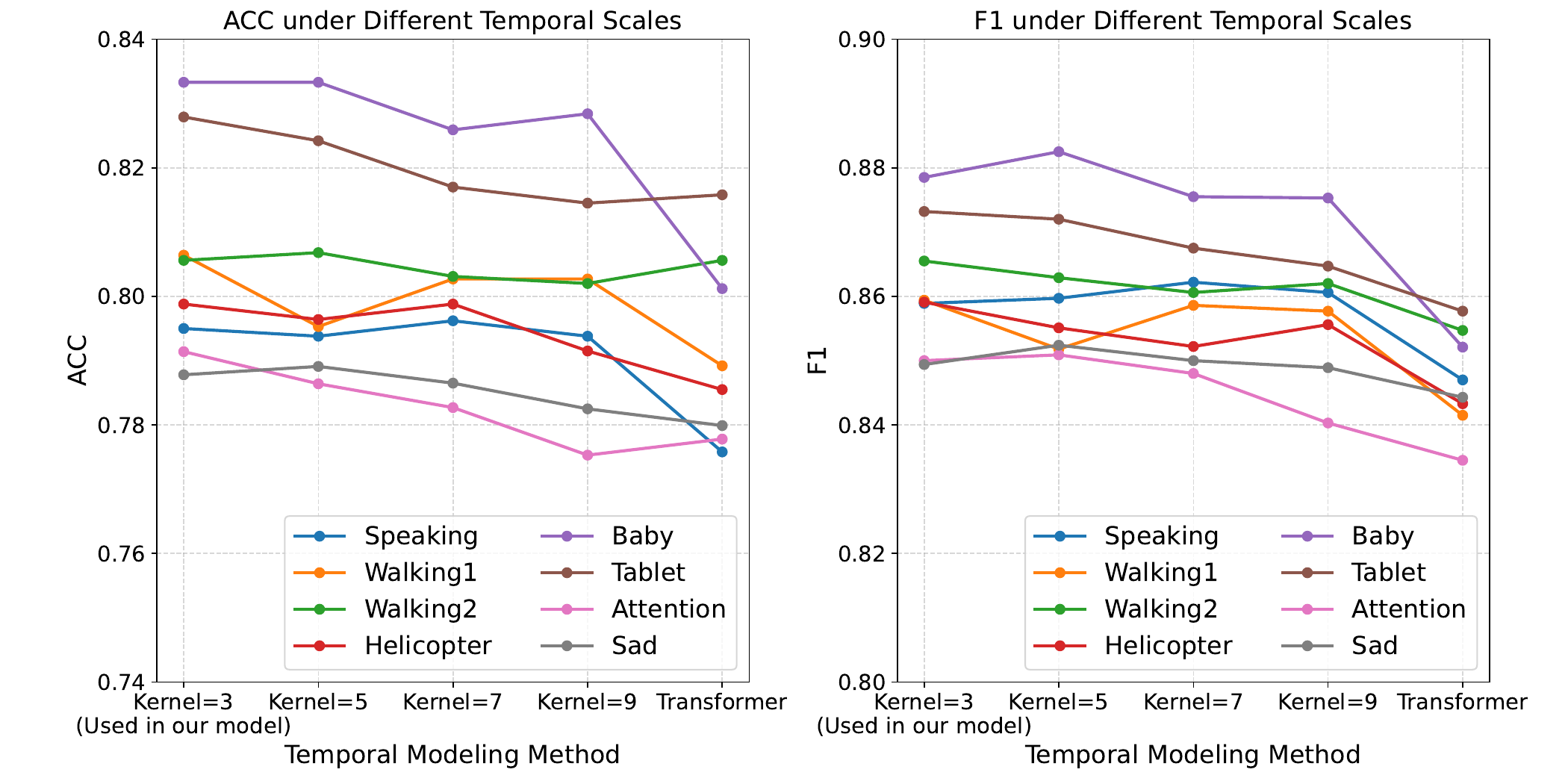}
    \caption{Comparison of different temporal modeling scales across datasets. }
    \label{fig:kernel_experiment}

\end{figure}

\subsection{Impact Analysis of Temporal Modeling Scales}

To empirically substantiate the foundational premise of emphasizing localized dynamics, we systematically evaluate the impact of different temporal modeling scales, directly addressing the utility of long-range modeling, based on the temporal branch only (i.e., Variant ${II}$ in the ablation study, which excludes the structural modeling component). First, we vary the kernel sizes within the Localized TCN branch ($k \in \{3, 5, 7, 9\}$) to modulate the receptive field. The results are summarized in Fig.~\ref{fig:kernel_experiment}. As shown in Fig.~\ref{fig:kernel_experiment}, we observe a continuous performance degradation as the kernel size within the localized temporal branch systematically expands from 3 to 9. This trend indicates that a compact receptive field ($k=3$) better preserves local fixation dynamics, whereas larger kernels may introduce redundant temporal context. To further investigate the impact of different temporal modeling strategies under short and sparsely sampled fixation sequences, we replace the TCN branch with a standard Multi-Head Self-Attention Transformer network. The Transformer-based variant achieves lower performance, which suggests that although global temporal dependency modeling can provide valuable information, directly applying unconstrained temporal aggregation may be less effective in capturing the structured characteristics of gaze behaviors in this setting. The results indicate that modeling localized event-driven temporal patterns, together with AOI-level relational structures, can provide more suitable representations for ASD-related gaze analysis.

\begin{figure}

    \centering
    \includegraphics[width=1\linewidth]{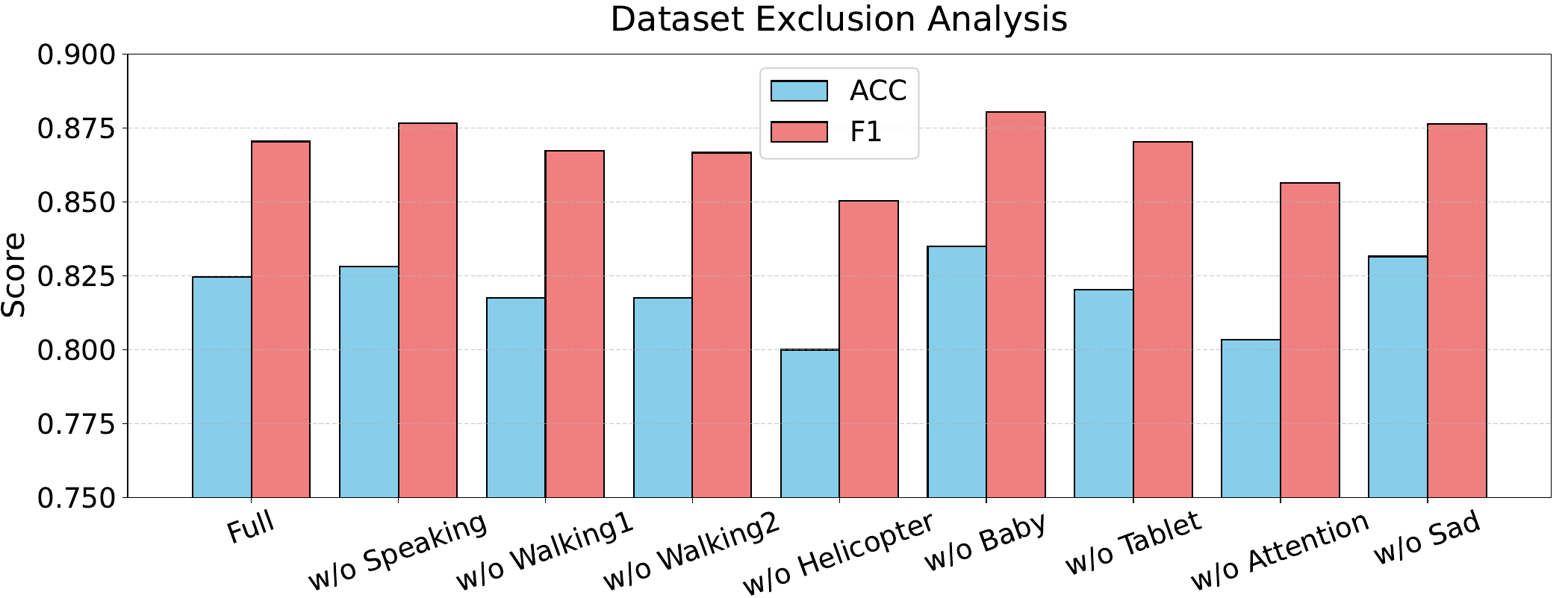}
    \caption{Generalization evaluation under missing stimulus paradigms. After a sample-level random split, the model is trained on samples containing all eight paradigms and evaluated on the test samples with one designated paradigm omitted.}
    \label{fig:placeholder}
\end{figure}

\subsection{Generalization Analysis under Dataset Exclusion}
To further investigate whether the proposed AOI-Net overly relies on specific dataset distributions or stimulus-dependent patterns, we conduct an additional cross-dataset generalization analysis. Specifically, we first construct a unified training setting using the complete collection of eye-movement datasets. To simulate a missing-paradigm scenario in deployment, all samples from a designated paradigm were deliberately removed from the test set (while retained in the training set), and the model was evaluated on the remaining test samples.
This setting aims to assess whether AOI-Net learns generalized ASD-related gaze representations that are shared across stimulus paradigms. Since real-world screening data may not always contain the full range of paradigms during model development, evaluating the model under dataset-exclusion settings provides a way to examine its robustness to variations in paradigm availability and data composition.
The corresponding results are illustrated in Fig.~\ref{fig:placeholder}. Interestingly, the exclusion of individual datasets does not consistently lead to performance degradation. While several datasets exhibit moderate performance drops, some datasets even achieve improved results compared with the full-data training setting. This phenomenon suggests that certain datasets may introduce distributional discrepancies or paradigm-specific variations that partially interfere with representation learning. Nevertheless, AOI-Net maintains stable overall performance across different dataset exclusion settings, demonstrating that the proposed AOI-guided structural modeling framework captures transferable and behaviorally meaningful gaze representations rather than overfitting to a single data source.
These findings further validate the robustness and cross-dataset generalization capability of AOI-Net under heterogeneous clinical conditions, supporting its applicability for real-world large-scale ASD screening scenarios.

\subsection{Discussion and Limitations}

AOI-Net is designed as a task-specific representation framework for ASD-related eye-tracking analysis, where fixation sequences and clinically meaningful AOI definitions provide behaviorally interpretable representations. Considering that the recorded gaze sequences are generally short and sparse under the current task setting, the temporal modeling strategy focuses on local fixation transitions and recurrent gaze patterns, which are frequently observed in ASD-related attentional behaviors. This design reflects the characteristics of the collected eye-tracking data and does not exclude the potential value of long-range dependencies. Future work will explore multi-scale temporal modeling to incorporate complementary global attention patterns.
The current evaluation is based on hospital cohorts, and potential differences may exist between clinical datasets and community screening populations. For example, community-based screening may involve different participant distributions, ASD prevalence, and recording conditions compared with hospital settings. Such variations could introduce distribution shifts that may influence model calibration and generalization. Future work will further evaluate the proposed framework on independent community cohorts and multi-site datasets to assess its robustness in broader screening scenarios.

\section{Conclusion}

In this work, we have proposed AOI-Net, an innovative gaze-behavior modeling framework that fundamentally advances the computational representation of eye-tracking data for ASD screening. Compared to the conventional solutions that predominantly rely on isolated temporal sequences or overlook the spatial semantics of AOIs, the proposed approach innovatively couples the short-term temporal dynamics of gaze events with the topological structures of semantically meaningful AOIs. To informatively and rationally fuse the heterogeneous behavioral signals, a novel adaptive network gating mechanism has been further introduced. Such a joint modeling strategy allows AOI-Net to capture the atypical attention allocation of ASD with both superior accuracy and explicit clinical interpretability. Unlike existing approaches evaluated under standardized experimental setups, the proposed framework is rigorously validated on a uniquely constructed, large-scale clinical database. Comprising eye-gaze track events from over 1,300 participants collected under eight distinct visual paradigms, the database provides a comprehensive resource for modeling complex behavioral traits. Furthermore, the complete framework and the standardized dataset have established a rigorous new benchmark for reproducible ASD research. Consequently, by shaping a robust computational paradigm for event-driven digital biomarker profiling, this research bridges the critical translational gap in AI-driven healthcare, providing a reliable foundation for deploying scalable neurodevelopmental screening tools in real-world clinical scenarios.

\bibliographystyle{IEEEtran}
\bibliography{refs}

\end{document}